\documentclass[10pt,twocolumn,letterpaper]{article}

\usepackage{iccv}
\usepackage{times}
\usepackage{epsfig}
\usepackage{graphicx}
\usepackage{amsmath}
\usepackage{amssymb}

\usepackage{multirow}
\usepackage[dvipsnames]{xcolor}
\usepackage{subcaption}
\usepackage[labelsep=period]{caption}
\usepackage{comment}
\DeclareMathAlphabet\mathbfcal{OMS}{cmsy}{b}{n}

\usepackage[linesnumbered,ruled,vlined]{algorithm2e}

\SetCommentSty{mycommfont}

\usepackage{adjustbox}
\usepackage{nicefrac}
\usepackage{booktabs}
\usepackage{tabularx}
\newcommand\setrow[1]{\gdef\rowmac{#1}#1\ignorespaces}
\newcommand\clearrow{\global\let\rowmac\relax}
\clearrow
\usepackage{pdfpages}

\usepackage[pagebackref=true,breaklinks=true,letterpaper=true,colorlinks,bookmarks=false]{hyperref}

\iccvfinalcopy 

\def\iccvPaperID{8360} 

\ificcvfinal\fi

\begin{document}

\title{Learning with Memory-based Virtual Classes for Deep Metric Learning}

\author{Byungsoo Ko\thanks{Authors contributed equally.}\textsuperscript{\rm 1}\\
NAVER/LINE Vision\\
{\tt\small kobiso62@gmail.com}
\and
Geonmo Gu\footnotemark[1]\textsuperscript{\rm 1}\\
NAVER/LINE Vision\\
{\tt\small korgm403@gmail.com}
\and
Han-Gyu Kim\\
NAVER Clova Speech\\
{\tt\small hangyu.kim@navercorp.com}
}

\maketitle

\begin{abstract}
The core of deep metric learning (DML) involves learning visual similarities in high-dimensional embedding space.
One of the main challenges is to generalize from seen classes of training data to unseen classes of test data.
Recent works have focused on exploiting past embeddings to increase the number of instances for the seen classes.
Such methods achieve performance improvement via augmentation, while the strong focus on seen classes still remains.
This can be undesirable for DML, where training and test data exhibit entirely different classes.
In this work, we present a novel training strategy for DML called MemVir.
Unlike previous works, MemVir memorizes both embedding features and class weights to utilize them as additional virtual classes.
The exploitation of virtual classes not only utilizes augmented information for training but also alleviates a strong focus on seen classes for better generalization.
Moreover, we embed the idea of curriculum learning by slowly adding virtual classes for a gradual increase in learning difficulty, which improves the learning stability as well as the final performance.
MemVir can be easily applied to many existing loss functions without any modification.
Extensive experimental results on famous benchmarks demonstrate the superiority of MemVir over state-of-the-art competitors.
Code of MemVir is publicly available\footnote{\url{https://github.com/navervision/MemVir}}.

\end{abstract}

\vspace{-5mm}
\section{Introduction}

\begin{figure}[t!]
\centering
\includegraphics[width=0.85\columnwidth]{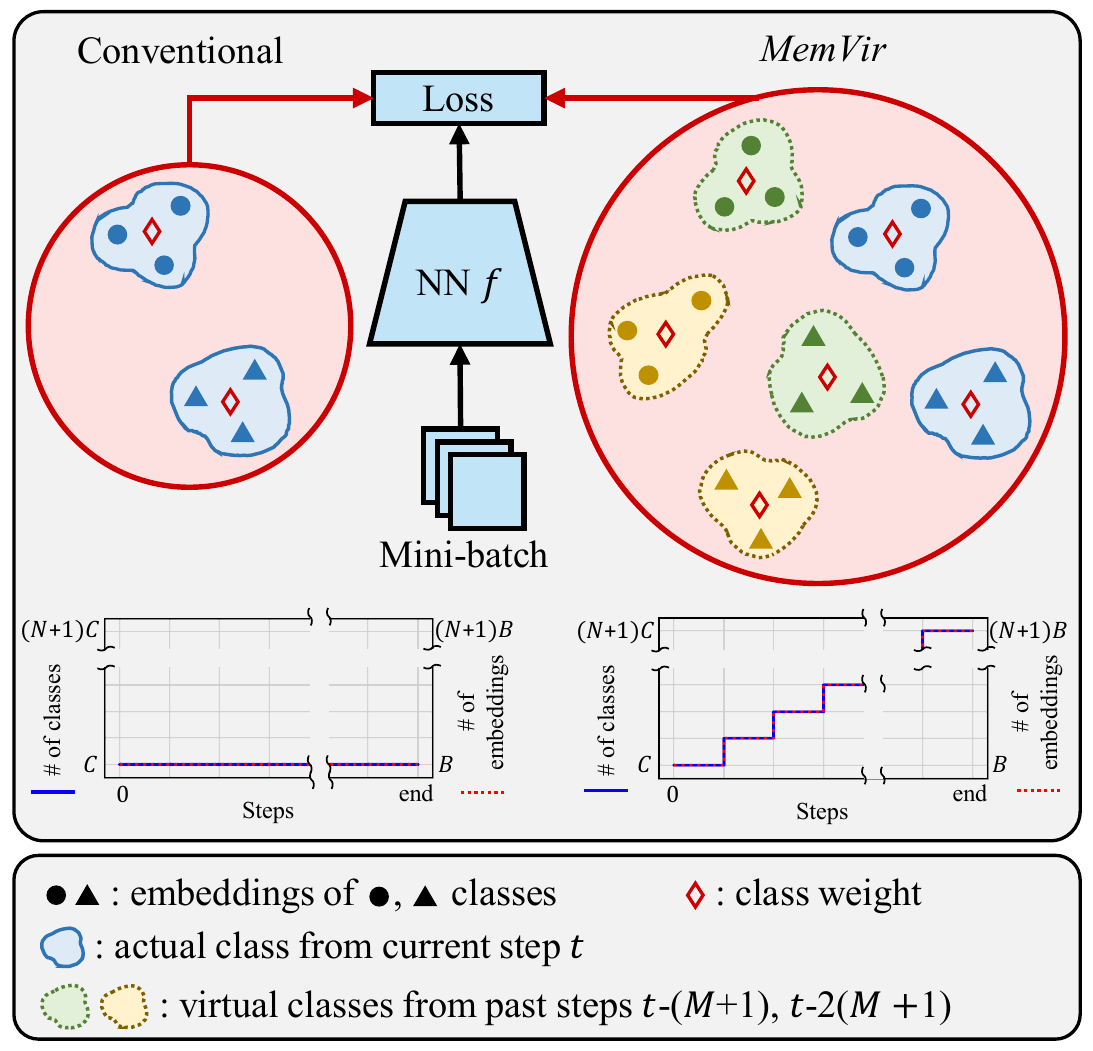}
\vspace{-0.5em}
\caption{In conventional training, the loss function is computed with actual classes. On the other hand, in MemVir, classes from previous steps (virtual classes) are used to compute the loss function along with the actual classes. Moreover, the number of classes and embeddings are gradually increased by adding virtual classes, where $C$ and $B$ denote number of classes and batch size, $N$ and $M$ are hyper-parameters for MemVir.\vspace{-1em}}
\label{fig:teaser}
\end{figure}

Deep metric learning (DML) is of great importance for learning visual similarities in a wide range of vision tasks, such as image clustering~\cite{hershey2016deep}, unsupervised learning~\cite{chen2020simple,he2020momentum,chen2020improved}, and image retrieval~\cite{wang2014learning,gordo2016deep,ko2020embedding,gu2020symmetrical}.
Learning visual similarity aims to build a well-generalized embedding space that reflects visual similarities of images using a defined distance metric.
Typically, training and test data exhibit entirely different classes in DML.
Thus, the main challenge is to maximize generalization performance from a training distribution to a shifted test distribution, which differs from classic classification tasks that deal with i.i.d. training and test distributions~\cite{milbich2020sharing,roth2020revisiting}.

Current DML approaches focus on learning visual similarities with objective functions, which considers pair-wise similarity (pair-based losses)~\cite{chopra2005learning,sohn2016improved,weinberger2009distance} or similarity between samples and class representatives (proxy-based losses)~\cite{wang2017normface, wang2018additive,liu2017sphereface,deng2019arcface,wang2018cosface}.
Recent studies propose exploiting additional embeddings from past training steps, which are saved and controlled in the memory queue, to increase the number of samples in a mini-batch and that of hard negative pairs~\cite{he2020momentum,chen2020improved,wang2020cross,kim2020broadface}.
And yet, these methods of utilizing past embeddings is still constrained to the seen classes of the training data.
Thus, the trained model might result to over-fit to the seen classes while under-perform on the unseen classes in test data.
Therefore, to learn an embedding space that generalizes, we need to alleviate the strong focus on seen classes during the training phase~\cite{roth2020revisiting,milbich2020sharing,milbich2020diva}.

In this paper, we propose a novel training strategy, which trains a model with Memory-based Virtual classes (MemVir), for DML.
In MemVir, we maintain memory queues for both class weights and embedding features.
Instead of using them to increase the number of instances of seen classes, they are treated as virtual classes to compute the loss function along with the actual classes, as illustrated in Figure~\ref{fig:teaser}.
Moreover, we incorporate the idea of curriculum learning (CL) to gradually increase the learning difficulty by slowly adding virtual classes.
The proposed MemVir has the following advantages:
\textbf{1)} MemVir trains a model with augmented information, which includes increased number of classes ($C \rightarrow (N+1)C$) and instances ($B \rightarrow (N+1)B$) without additional feature extraction.
\textbf{2)} CL-like gradually increasing the learning difficulty improves the optimization stability and final performance.
\textbf{3)} Exploiting virtual classes help achieve more generalized embedding space by alleviating excessively strong focus on seen classes of training data.
\textbf{4)} MemVir can be easily applied to many existing loss functions to obtain a significant performance boost without any modification of the loss function.

\textbf{Contributions. }
\textbf{1)} We propose a novel training strategy for DML that exploits past embeddings and class weights as virtual classes to improve generalization.
We further improve the training process and performance by incorporating the idea of CL.
\textbf{2)} We exhaustively analyze our proposed method and demonstrate that employing virtual classes improves generalization by alleviating a strong focus on seen classes theoretically and empirically.
\textbf{3)} MemVir achieves state-of-the-art performance on three popular benchmarks of DML in both conventional and \textit{Metric Learning Reality Check (MLRC)}~\cite{musgrave2020metric} evaluation protocol.

\section{Related Work}

\textbf{Sample Generation and Memory-based Learning. }
In DML, the generation of hard samples has been investigated to perform training with more informative samples~\cite{duan2018deep,zheng2019hardness,gu2020symmetrical,ko2020embedding}.
DAML~\cite{duan2018deep} and HDML~\cite{zheng2019hardness} utilize generative networks to generate synthetic samples, while Symm~\cite{gu2020symmetrical} and EE~\cite{ko2020embedding} generate synthetic samples by geometric relations.
Meanwhile, utilizing information from previous steps has been explored in many computer vision tasks~\cite{he2020momentum,chen2020improved,wang2020cross,kim2020broadface}.
In supervised DML, XBM~\cite{wang2020cross} is proposed to use memorized embeddings for extending negative samples in pair-based losses.
In XBM, the state difference between past and current embeddings is disregarded based on ``\textit{slow drift}'' phenomena.
On the other hand, \cite{kim2020broadface} argues that a large accumulated error caused by the state difference may degrade the training process.
They present BroadFace method for softmax variant losses to control the error by compensating the state difference and gradient control.
The above-mentioned methods focus on utilizing generated or memorized information with respect to increasing the number of instances for the seen classes.
However, this may result in a model overly optimized to the seen classes while under-performing on the unseen classes in test data.
Rather than disregarding or controlling the state difference, the proposed MemVir exploits the state difference by employing the memorized embeddings and class weights as \textit{virtual classes}, which are treated as different classes from the actual (seen) classes.
The exploitation of virtual classes helps achieve more generalized embedding space by alleviating a strong focus on seen classes.
Additional comparison with XBM w.r.t ``\textit{slow drift}'' phenomena is included in supplementary Section B.2.

\textbf{Virtual Class. }
In image recognition task, Virtual softmax~\cite{chen2018virtual} has been presented to enhance the discriminative property of embeddings by injecting a virtual class into the softmax loss.
However, it is not only limited by a single virtual class but also cannot be used with softmax variants using $l_2$-normalization.
In comparison, MemVir exploits multiple virtual classes and can be used with any softmax variants and proxy-based losses.

\textbf{Curriculum Learning. }
CL in machine learning is motivated by the idea of curriculum in human learning, where the models learn from easier samples first and more difficult samples later.
Imposing CL for model training has been shown to accelerate and improve the training process in many machine learning tasks~\cite{bengio2009curriculum, weinshall2018curriculum, hacohen2019power,huang2020curricularface}.
When exploiting CL, two key factors have to be considered:
(1) Scoring the difficulty of each sample; (2) scheduling the pace by which the sample is presented to the network.
To define the difficulty, bootstrapping and transfer learning have been used to score the difficulty of each sample~\cite{weinshall2018curriculum,hacohen2019power}.
For scheduling, the samples to be presented to the network can be determined in fixed or adaptive steps~\cite{weinshall2018curriculum,huang2020curricularface}.
The main difference between conventional CL and MemVir is the former schedules within the training data, whereas the latter (MemVir) increases the learning difficulty with virtual classes, which are augmented information.

\begin{figure*}[t!]
\centering
\vspace{-2mm}
\includegraphics[width=0.73\textwidth]{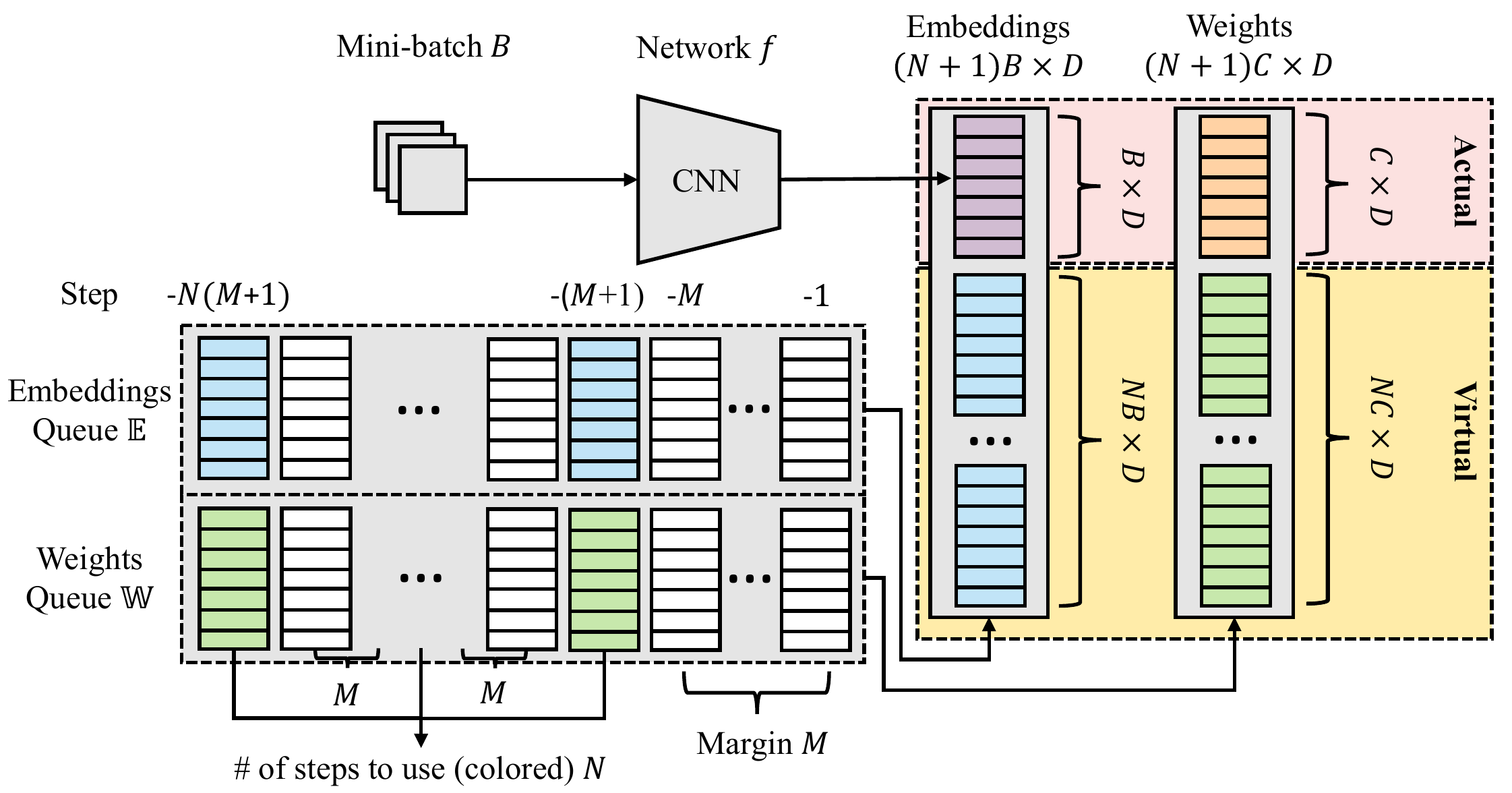}
\vspace{-2mm}
\caption{Overview of MemVir. Past embeddings and class weights queues are maintained. We select $N$ steps of past embeddings and weights with margin $M$ in between the selected steps, and use them as additional virtual classes along with actual classes for loss computation.\vspace{-3mm}}
\label{fig:framework}
\end{figure*}

\section{Proposed Method}

\subsection{Preliminary}

We define a deep neural network as $f:\mathbfcal{I} \rightarrow \mathbfcal{X}$, which is a mapping from an input data space $\mathbfcal{I}$ to an embedding space $\mathbfcal{X}$.
Let $X = [x_1, x_2, \dots, x_H]$ denote the $D$-dimensional embedding features, and each feature $x_i$ has a corresponding label $y_i \in \{1, \dots ,C\}$.
The generalized form of the objective function can be written as follows:
\begin{eqnarray}
\mathcal{L}(X,W) = -\frac{1}{\mid X \mid}\sum_{i=1}^{\mid X \mid} l(x_i, y_i),
\end{eqnarray}
where $W$ denotes the class weights, and $l(\cdot)$ can be any of the loss functions defined below.

The most widely used classification loss function, softmax loss, has been revalued as a competitive objective function in metric learning~\cite{zhai2018classification, boudiaf2020metric}.
The softmax loss is used to optimize the network $f$ and class weight $W$:
\begin{eqnarray}
l_{softmax}(x_i,y_i) = \log\frac{e^{W^T_{y_i}x_i}}{\sum_{j=1}^{C}e^{W^T_j x_i}},
\label{eq:softmax}
\end{eqnarray}
where $W_j \in \mathbb{R}^D$ denotes the $j$-th column of weight $W \in \mathbb{R}^{D \times C}$.
The bias $b$ is set to $0$ because it does not affect the performance~\cite{liu2017sphereface, deng2019arcface}.
The weight $W_j$ is the center of each class~\cite{deng2019arcface, wang2018cosface} and serves as a representative.

For improved performance and better interpretation, \cite{wang2017normface, wang2018additive, liu2017sphereface} proposes to normalize weights and embedding features to lay them on a hypersphere with a fixed radius.
We perform $l_2$-normalization to fix the size of the weights and embedding features to the following: $\parallel W_j \parallel = 1$ and feature $\parallel x_i \parallel = 1$.
Subsequently, we can simplify the logits~\cite{pereyra2017regularizing} by transforming $W_j^T x_i = {\parallel W_j \parallel \parallel x_i \parallel \cos{\theta_j}} = \cos{\theta_j}$, and define the Norm-softmax loss as follows:
\begin{eqnarray}
l_{norm}(x_i,y_i) = \log\frac{e^{\gamma \cos{\theta_{y_i}}}}{e^{\gamma \cos{\theta_{y_i}}} + \sum_{j=1, j\neq y_i}^{C}e^{\gamma \cos{\theta_{j}}}},
\end{eqnarray}
where $\gamma$ is a scale factor.
The proposed method MemVir can be used with softmax variants as well as proxy-based losses because a proxy is a class representative feature much like class weights of softmax variants.
Hence, we include the details of other loss functions (CosFace~\cite{wang2018cosface}, ArcFace~\cite{deng2019arcface}, CurricularFace~\cite{huang2020curricularface}, Proxy-NCA~\cite{movshovitz2017no}, and Proxy-Anchor~\cite{kim2020proxy}) in supplementary Section A.

\subsection{Learning with Memory-based Virtual Classes}

We propose a novel training strategy called MemVir, which trains a model with virtual classes from past steps to exploit augmented information and obtain better generalization.
When conventional metric learning trains a model with given $C$ classes and $B$ embeddings from training data, MemVir gradually increases the number of classes ($C \rightarrow (N+1)C$) and embeddings ($B \rightarrow (N+1)B$) with the virtual classes.
We use the naming convention of MemVir($N$,$M$), which indicates the hyper-parameters of the proposed method, to be defined below.

\vspace{-3mm}
\paragraph{Queuing Past Embeddings and Weights.}
To form a class in loss computation, a pair of the class representative feature (weight) and embedding features are required.
Hence, in MemVir, we maintain two types of memory queues: embedding queue $\mathbb{E}$ and weight queue $\mathbb{W}$, where each entity of the queues is a collection of embeddings or class weights of each step as illustrated in Figure~\ref{fig:framework}.
For each step, the collection of embeddings $X$ and weights $W$ are enqueued to $\mathbb{E}$ and $\mathbb{W}$, respectively.
The size of each queue is determined as $N(M+1)$, where $N$ is the number of selected steps to use for the loss computation, and $M$ is the margin between the selected steps.
The shape and position of class clusters vary by each step because the network parameters change during training process.
Such variance between steps is utilized in MemVir by exploiting weights and embeddings from previous steps as virtual classes.
Here, the difference between the selected steps can be controlled by the margin $M$.

\setlength{\textfloatsep}{5pt}
\begin{algorithm}[t!]
\footnotesize
\DontPrintSemicolon
\SetAlgoLined
\tcp{$f$: encoder network}
\tcp{weight$/$embed\_queue: weight and embedding memory queue}
\tcp{$U_e$, $N$, $M$: warm-up epoch, number of steps, margin}

\For{input, label in loader}{
    embed = $f$.forward(input)\\
    weight = $f$.get\_class\_weight()\\
    
    \tcp{Turn on MemVir when it is in use and past warm-up epoch}
    \uIf{MemVir is True and epoch $\geq U_e$}{
        cur\_weight = weight.copy() \\
        cur\_embed = embed.copy() \\
        cur\_label = label.copy() \\
        \tcp{Prepare embeddings and weights by step-pacing}
        \tcp{The order of each queue is from new to old}
        \uIf{len(weight\_queue) $> M$}{
        \For{idx in range($M$, len(weight\_queue), $M+1$)}{
            pre\_weight = weight\_queue[idx] \\
            pre\_embed, pre\_label = embed\_queue[idx] \\
            \tcp{Create new label indices for virtual classes}
            new\_label = create\_new\_label(pre\_label) \\
            weight.concatenate(weight, pre\_weight) \\
            embed.concatenate(embed, pre\_embed) \\
            label.concatenate(label, new\_label) \\
        }}

        \tcp{Update memory queues}
        enqueue(weight\_queue, cur\_weight)\\
        enqueue(embed\_queue, (cur\_embed, cur\_label))\\
        \uIf{len(weight\_queue) $>N(M+1)$}{
            dequeue(weight\_queue)\\
            dequeue(embed\_queue)\\
        }
    }

    \tcp{Compute loss and back-propagation}
    loss = compute\_loss(weight, embed, label)\\
    loss.backward()\\
    optimizer.step()
}
\caption{Pseudo-code of MemVir}
\label{algo:code}
\end{algorithm}
\setlength{\textfloatsep}{15pt}

\begin{figure}[t!]
\centering
     \begin{subfigure}[b]{0.495\columnwidth}
         \centering
         \includegraphics[width=1.0\columnwidth]{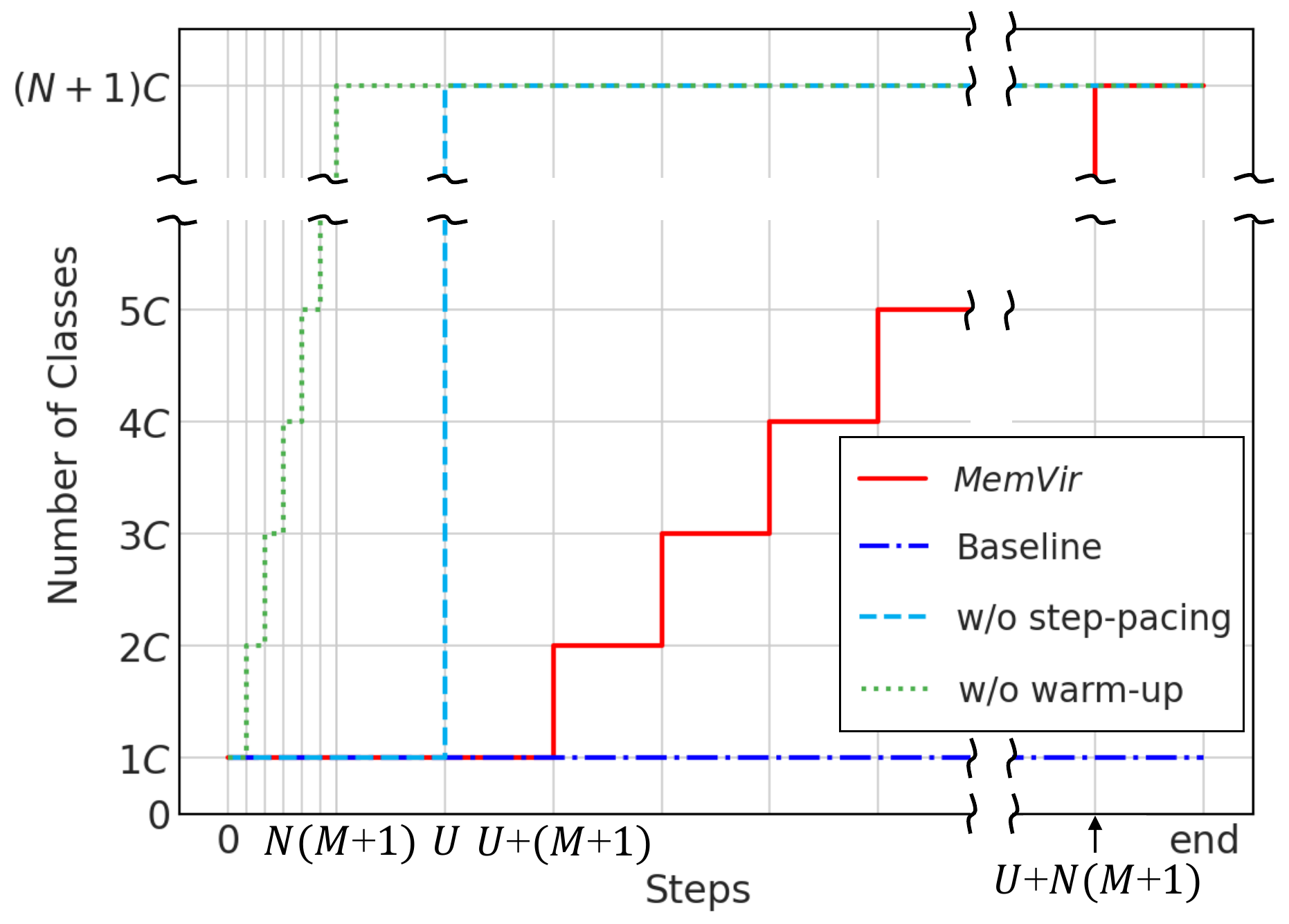}
         \caption{Different ways of scheduling.}
         \label{fig:schedule_class}
     \end{subfigure}
     \begin{subfigure}[b]{0.495\columnwidth}
         \centering
         \includegraphics[width=1.0\columnwidth]{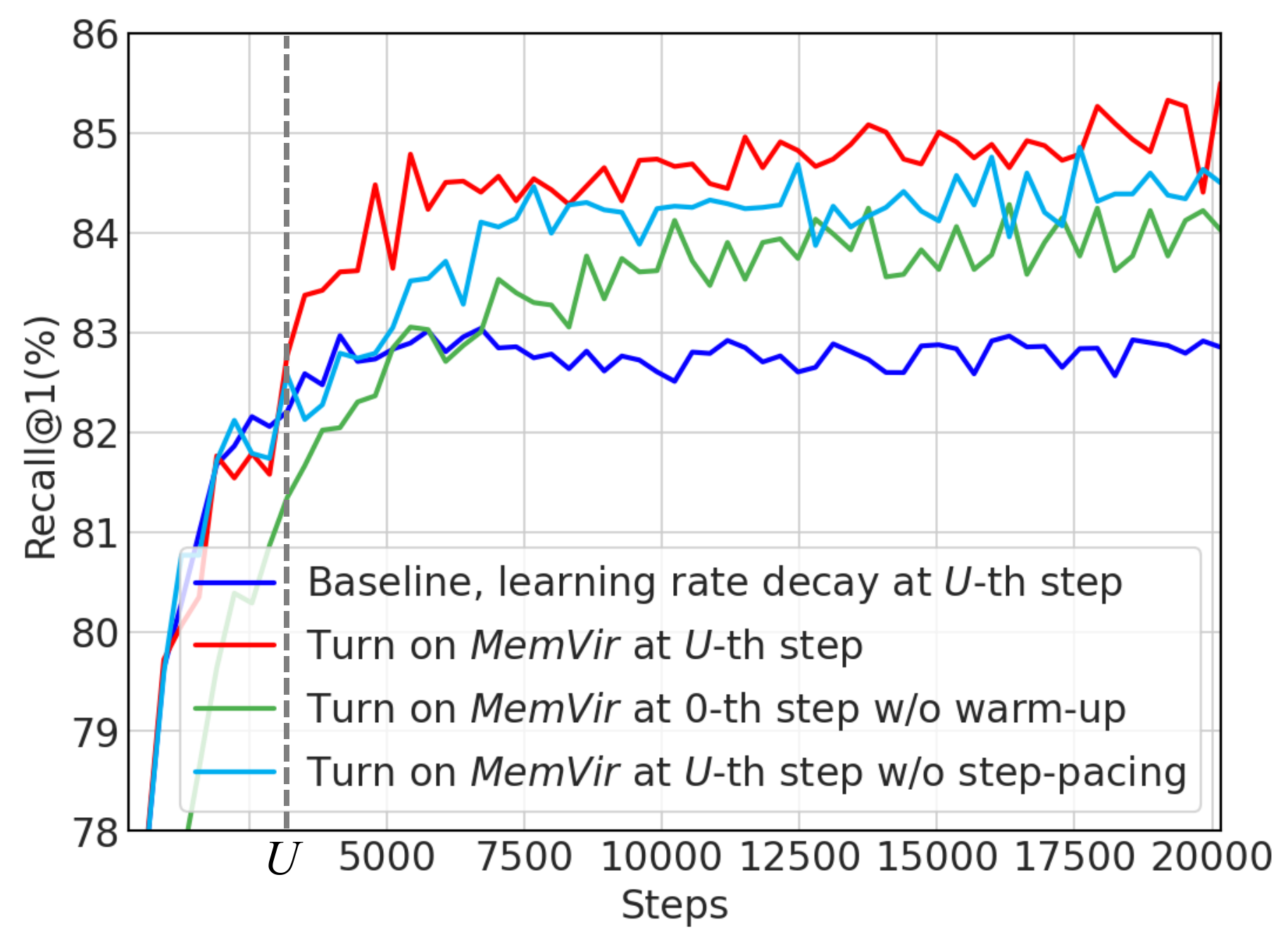}
         \caption{Performance by scheduling.}
         \label{fig:schedule_acc}
     \end{subfigure}
\caption{Impact of scheduling. (a) Different ways of scheduling for adding virtual classes. (b) Performance of each scheduling case with MemVir(5,100) and Norm-softmax as baseline on CARS196.}
\label{fig:schedule}
\end{figure}

\vspace{-3mm}
\paragraph{Scheduling Usage of Virtual Classes.}
In MemVir, virtual classes will be utilized to gradually increase learning difficulty as CL.
The scheduling of virtual class usage includes two periods: \textit{warm-up} and \textit{step-pacing}.
We turn on MemVir and begin managing queues after the warm-up step $U$ (epoch $U_e$), because the embeddings of the initial phase are typically scattered without forming clusters, which can be a distraction for training.
It is noteworthy that we use MemVir without learning rate decay because decaying the learning rate changes the difference between steps; thus, the learning rate decay can be used with a modification of hyper-parameter $M$ of MemVir.
After the warm-up, the step-pacing algorithm takes place by storing embeddings and weights of each step in their respective queues and reusing them for loss computations, as described in Algorithm~\ref{algo:code}.
As the queue size grows, previously stored embeddings and weights from every $M+1$ steps are selected as virtual classes when computing the loss at each step.
The number of selected steps for virtual classes would increase gradually from 0 to $N$ determined by current queue size.
This results in increasing the number of classes by a staircase function, and the function $s$ of the number of classes can be written as:
\begin{eqnarray}
s(i)=
\begin{cases}
C,  &i<U, \\
C\times\left\{min(\lfloor \frac{i-U}{M+1} \rfloor,N)+1\right\}, &i\geq U,
\end{cases}
\end{eqnarray}
where $i$ denotes the current step.
The scheduling function of MemVir is illustrated by the red line in Figure~\ref{fig:schedule_class}.

\paragraph{Learning with Multiple Virtual Classes.}
When we select $N$ steps of past embeddings and weights from the queues, it indicates that we have $NC$ virtual classes.
We denote the set of selected past embeddings and weights as $\widetilde{X}$ and $\widetilde{W}$, respectively.
Subsequently, we compute the objective function with virtual classes as follows:
\begin{eqnarray}
\mathcal{L}(X\cup\widetilde{X},W\cup\widetilde{W}) = -\frac{1}{\mid X\cup\widetilde{X} \mid}\sum_{i=1}^{\mid X\cup\widetilde{X} \mid} l(x_i, y_i),
\end{eqnarray}
where $l(\cdot)$ can be any type of loss function.
The implementation of MemVir is simple without any modification of the loss function, and it gives a significant performance improvement in DML without any additional computational cost in the inference phase.

\subsection{Discussion and Analysis}

\subsubsection{Analysis of Scheduling}
Figure~\ref{fig:schedule} shows the different ways of scheduling and the performance of each case.
In Figure~\ref{fig:schedule_class}, when the MemVir is turned on at warm-up step $U$, it begins adding virtual classes after each $M+1$ step, gradually.
Compared with MemVir, `w/o warm-up' starts adding virtual classes from the initial steps, whereas `w/o step-pacing' adds all virtual classes at once after warm-up step $U$.
For the case of `w/o warm-up', training starts with degraded performance, but finally, the performance is higher than the baseline.
In fact, embeddings from virtual classes at the initial steps would be scattered without forming clusters; thus, it can be a distraction at the initial steps.
Meanwhile, `w/o step-pacing' exhibits a slight performance degradation immediately after warm-up step $U$.
This is because placing $NC$ number of virtual classes simultaneously can be too difficult for training the model.
By considering both cases, MemVir is able to increase the training difficulty gradually for a more stable optimization.

\subsubsection{Analysis of Difficulty}
\begin{figure}[t!]
    \centering
     \begin{subfigure}[b]{0.5\columnwidth}
        \centering\includegraphics[width=1.0\columnwidth]{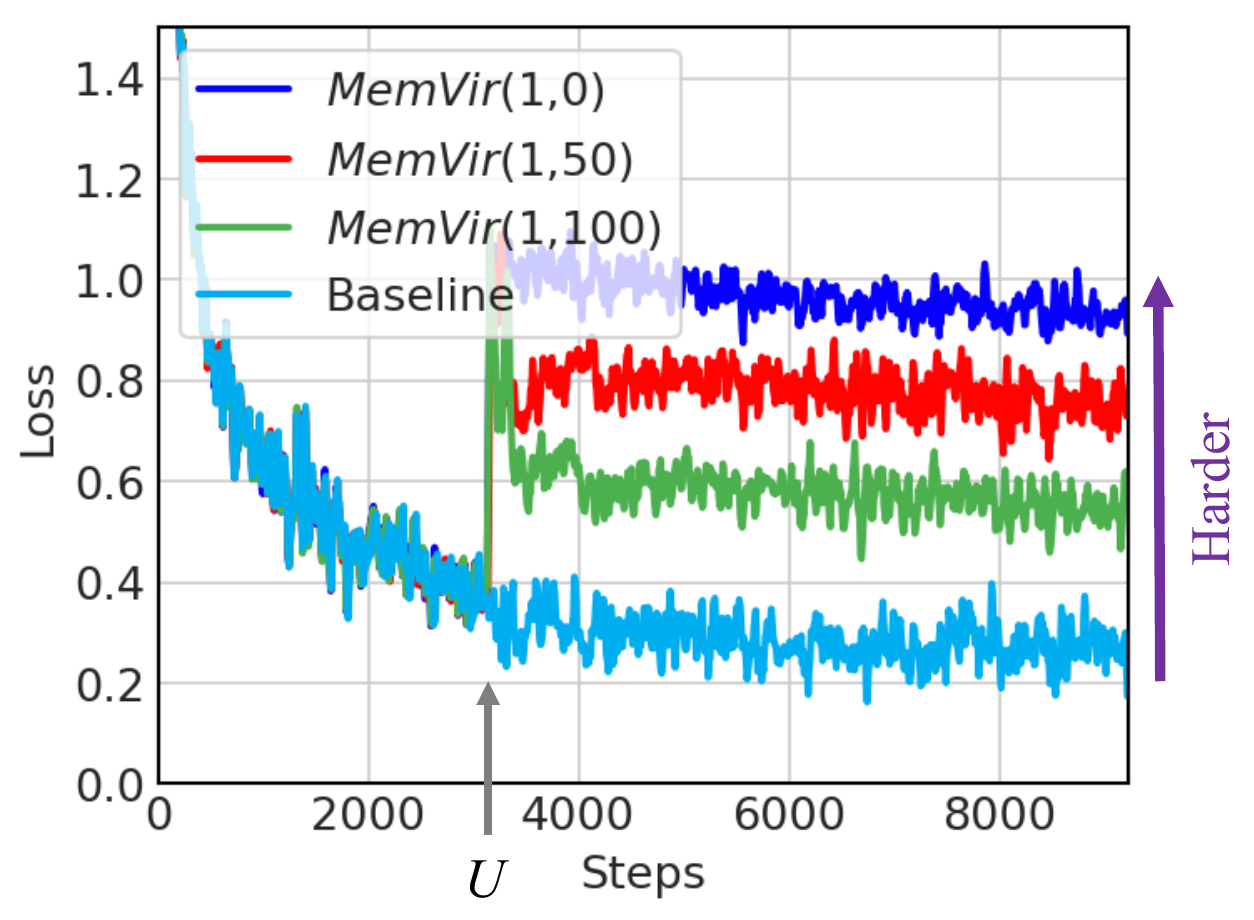}
         \caption{Difficulty by margin.}
         \label{fig:difficulty_m}
     \end{subfigure}\hfill
    \begin{subfigure}[b]{0.5\columnwidth}
        \centering\includegraphics[width=1.0\columnwidth]{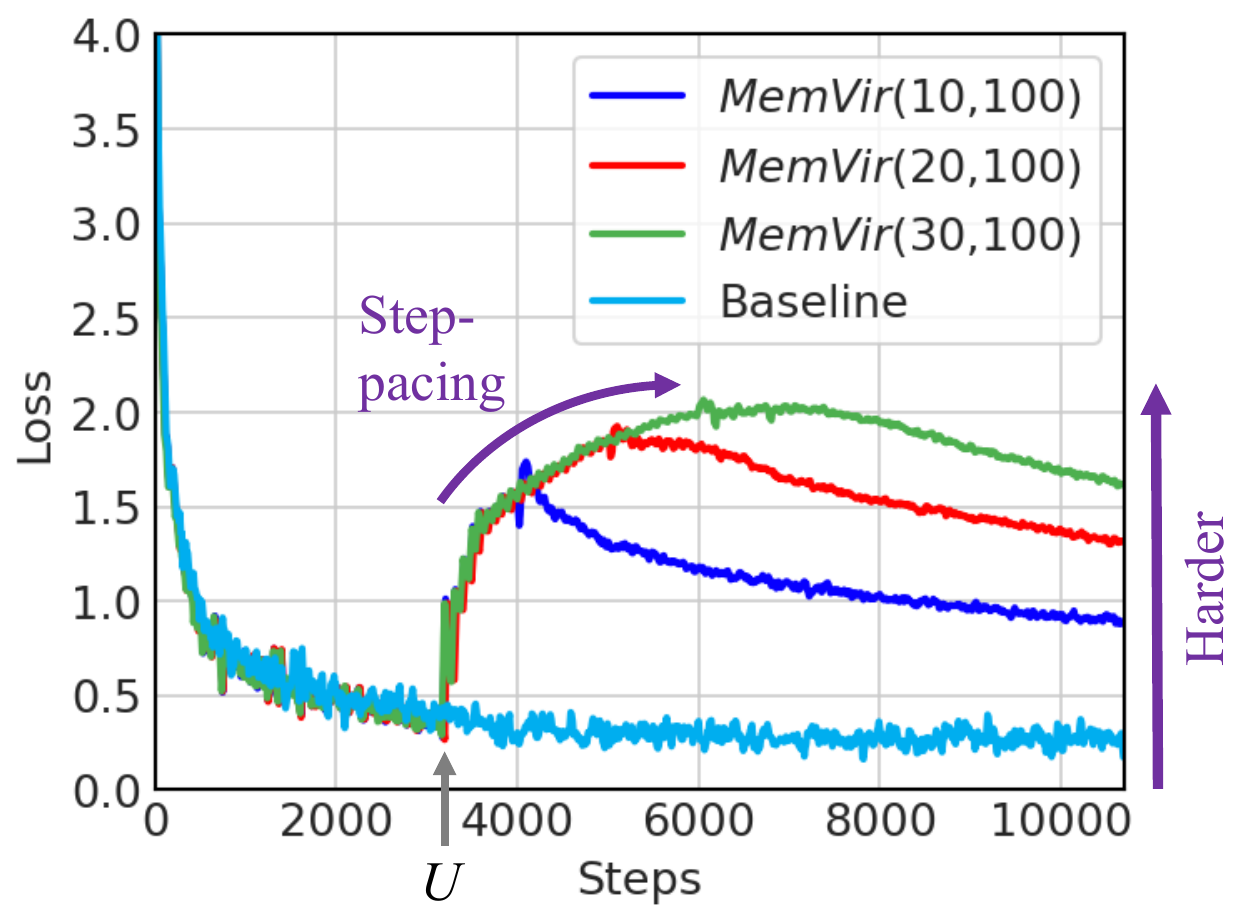}
         \caption{Difficulty by number of steps.}
         \label{fig:difficulty_n}
     \end{subfigure}
\caption{Impact of difficulty with Norm-softmax as baseline on CARS196. Difficulty is measured by loss value of each step. (a) Difficulty by varying margin parameter $M$ with a fixed number of steps $N=1$. (b) Difficulty by varying the number of steps $N$ with a fixed margin of $M=100$.}
\label{fig:difficulty}
\end{figure}

MemVir controls learning difficulty via following hyper-parameters: number of steps $N$ and margin $M$.
To see the impact of learning difficulty by each hyper-parameter, we measure the difficulty with the loss value by following~\cite{kumar2010self,weinshall2018curriculum}.
As shown in Figure~\ref{fig:difficulty_m}, a smaller margin of $M$ results in greater difficulty, which is obvious because the embeddings from the recent steps would be similar to the embeddings from the current steps.
Furthermore, Figure~\ref{fig:difficulty_n} shows that adding more virtual classes increases the learning difficulty.
It is noteworthy that the loss value increases slowly after warm-up step $U$ by adding virtual classes gradually (step-pacing); subsequently, it starts decreasing after reaching a peak.
The detailed performance by different hyper-parameters is presented in Section~\ref{sec:param}.

\subsubsection{Gradient Analysis for Generalization}
Considering the distribution shift in training and test data, strong focus on seen classes has to be alleviated in the generalization of transfer learning problems such as DML~\cite{roth2020revisiting,milbich2020sharing,milbich2020diva}.
To demonstrate how MemVir works in generalizing models during training, we have analyzed the gradient of the softmax loss.
For convenient analysis, the softmax loss in Equation~\ref{eq:softmax} is re-written as follows:
\begin{eqnarray}
l_{softmax}(x_i, y_i) & = & \log \frac{e^{\alpha(x_i, y_i)}}{\sum_{j=1}^C e^{\alpha(x_i, j)}},
\end{eqnarray}
where $\alpha(x_i, j) = W_{j}^T x_i$.         
The gradient of the softmax loss over the embedding feature $x_i$ can be inducted as follows:
\begin{eqnarray}
\frac{\partial l_{softmax}(x_i, y_i)}{\partial x_i} & = & W_{y_i} - \frac{\sum_{j=1}^C e^{\alpha(x_i, j)} W_j}{\sum_{j=1}^C e^{\alpha(x_i, j)}} \nonumber \\
        & \approx & W_{y_i} - \frac{e^{\alpha(x_i, y_i)} W_{y_i}}{\sum_{j=1}^C e^{\alpha(x_i, j)}} \nonumber \\
        & = & \tau W_{y_i},
\end{eqnarray}
\begin{eqnarray}
\tau & = & 1 - \frac{e^{\alpha(x_i, y_i)}}{\sum_{j=1}^C e^{\alpha(x_i, j)}}.
\end{eqnarray}
It is obvious that $\tau > 0$ and $\tau \rightarrow{} 0$ when $x_i \rightarrow{} W_{y_i}$, implying that $x_i$ tries to get as close to $W_{y_i}$ as possible, which is illustrated in Figure~\ref{fig:proof_softmax}.
This can result in a strong focus on the target weight $W_{y_i}$ and an over-fit to the seen classes of the training data.

\begin{figure}[t!]
    \centering
     \begin{subfigure}[b]{0.5\columnwidth}
        \centering\includegraphics[width=1.0\columnwidth]{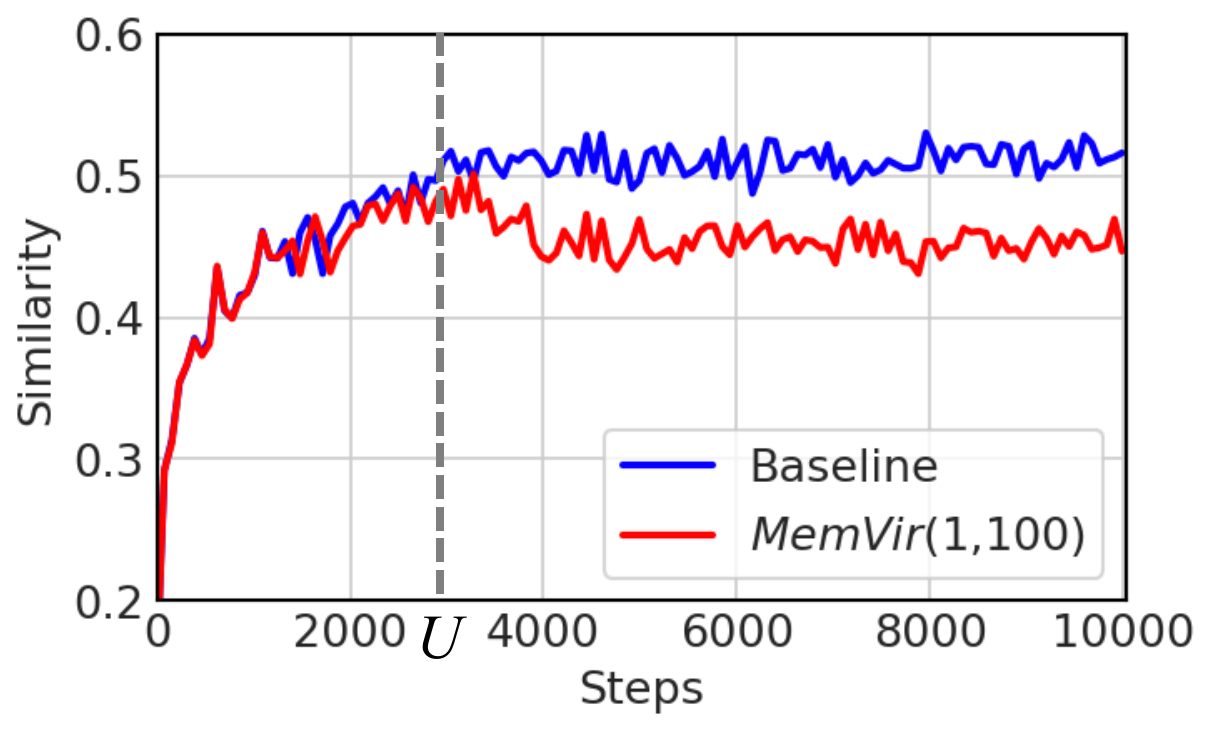}
         \caption{Cosine similarity ($x_i$, $W_{y_i}$).}
         \label{fig:gen_sim}
     \end{subfigure}\hfill
    \begin{subfigure}[b]{0.5\columnwidth}
        \centering\includegraphics[width=1.0\columnwidth]{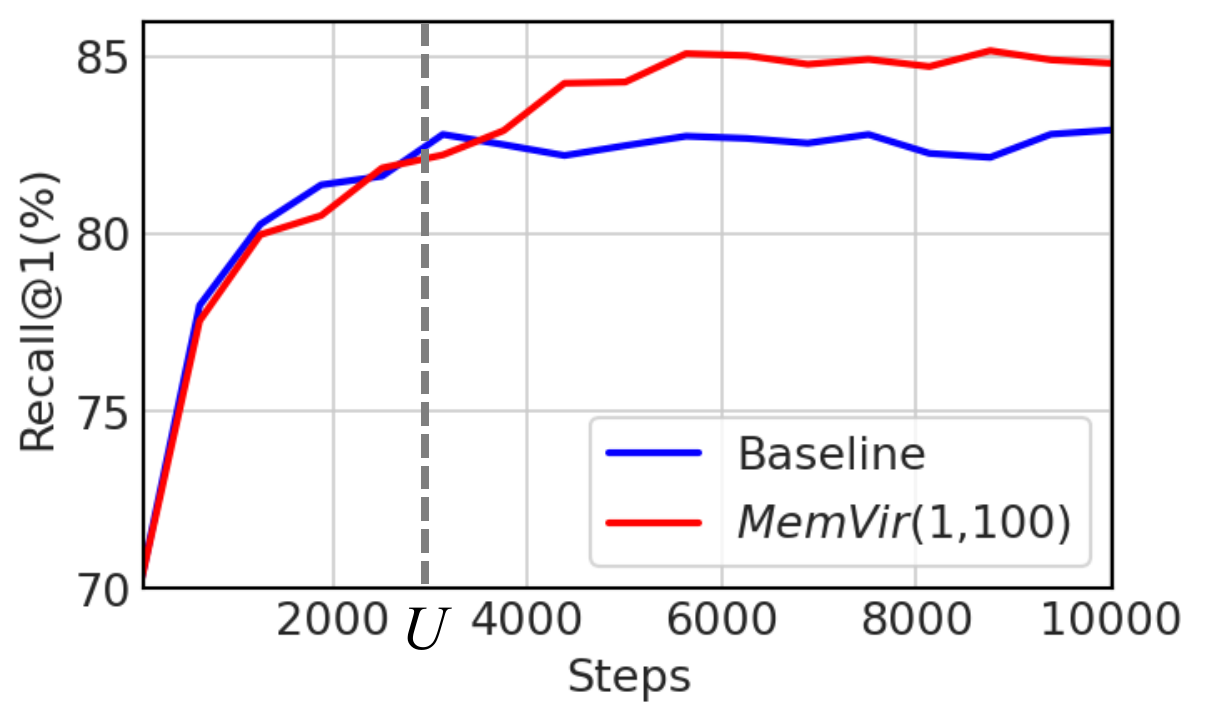}
         \caption{Generalization performance.}
         \label{fig:gen_acc}
     \end{subfigure}
\caption{Generalization analysis with Norm-softmax as baseline on CARS196. (a) Similarity between embeddings and corresponding class weights of seen classes in training data. (b) Performance on unseen classes in test data.}
\label{fig:generalization}
\end{figure}

\begin{figure}[t!]
    \centering
     \begin{subfigure}[b]{0.5\columnwidth}
        \centering\includegraphics[width=0.75\columnwidth]{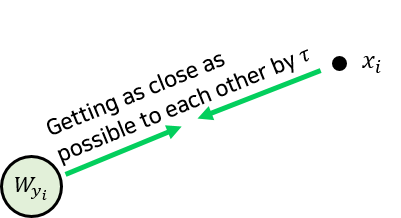}
         \caption{Softmax loss.}
         \label{fig:proof_softmax}
     \end{subfigure}\hfill
    \begin{subfigure}[b]{0.5\columnwidth}
        \centering\includegraphics[width=0.75\columnwidth]{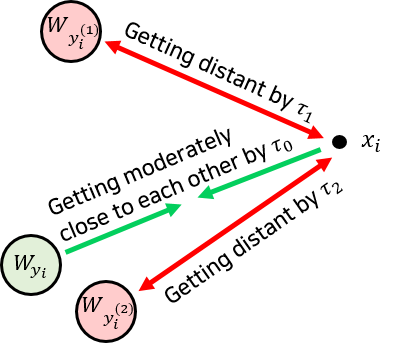}
         \caption{MemVir + softmax loss.}
         \label{fig:proof_memvir}
     \end{subfigure}
\caption{Illustration of an embedding ($x_i$) and corresponding class weight ($W_{y_i}$) learning, where $W_{{y_i}^{(n)}}$ are virtual class weights originated from the class $y_i$.}
\label{fig:proof}
\end{figure}

\begin{figure*}[t!h!]
\vspace{-2mm}
     \centering
     \begin{subfigure}[b]{0.8\linewidth}
         \centering
         \includegraphics[width=1.0\columnwidth]{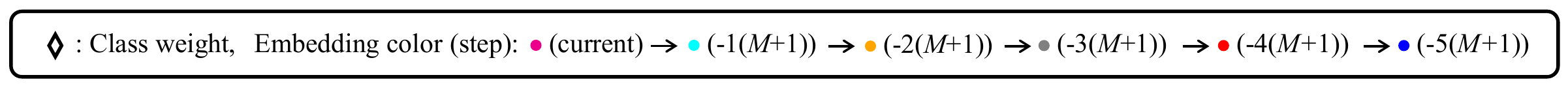}
     \end{subfigure}
     \begin{subfigure}[b]{0.32\linewidth}
         \centering
         \includegraphics[width=1.0\columnwidth]{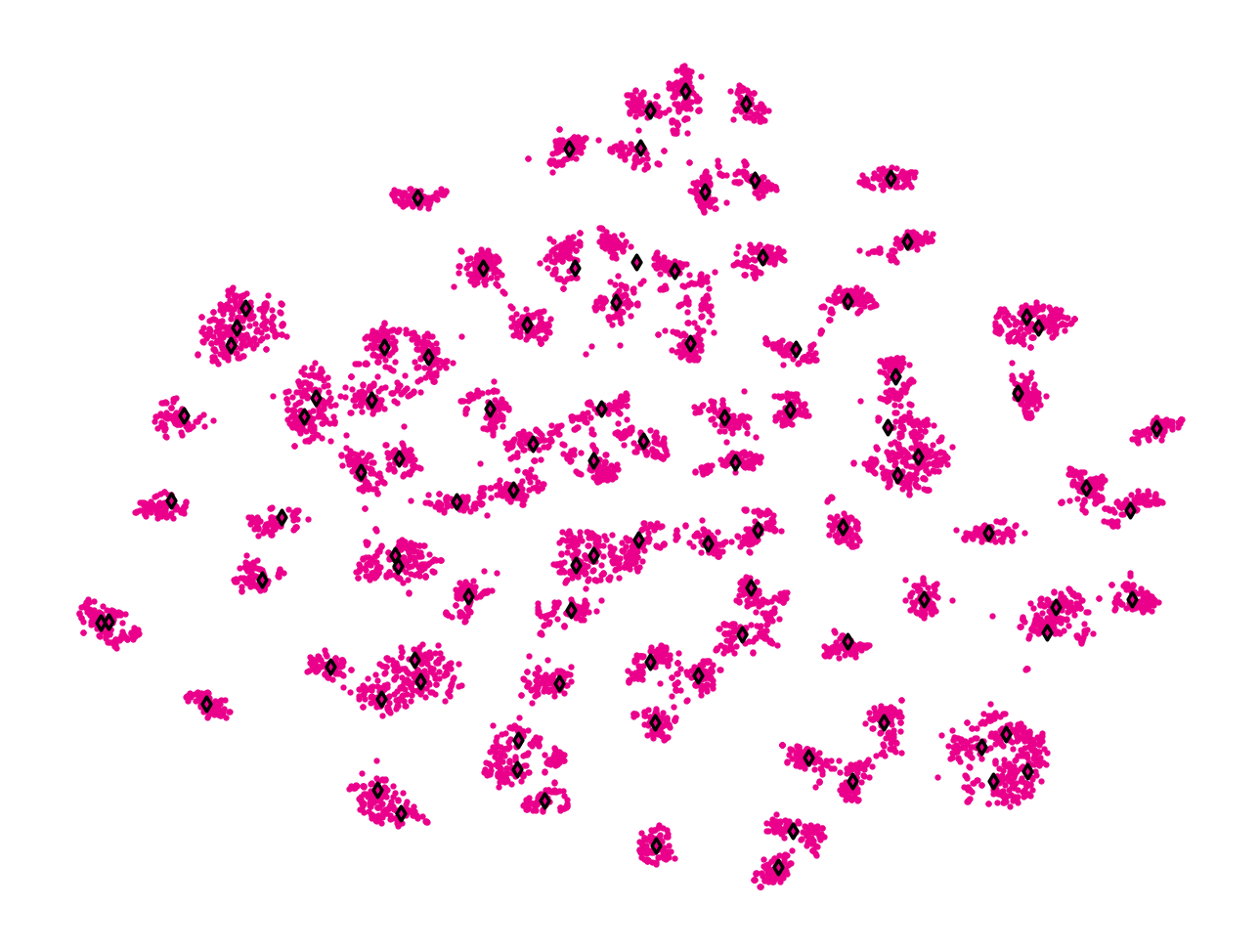}
         \caption{50th epoch, $\# \:of\: classes=C$}
         \label{fig:vis_b}
     \end{subfigure}
     \begin{subfigure}[b]{0.32\linewidth}
         \centering
         \includegraphics[width=1.0\columnwidth]{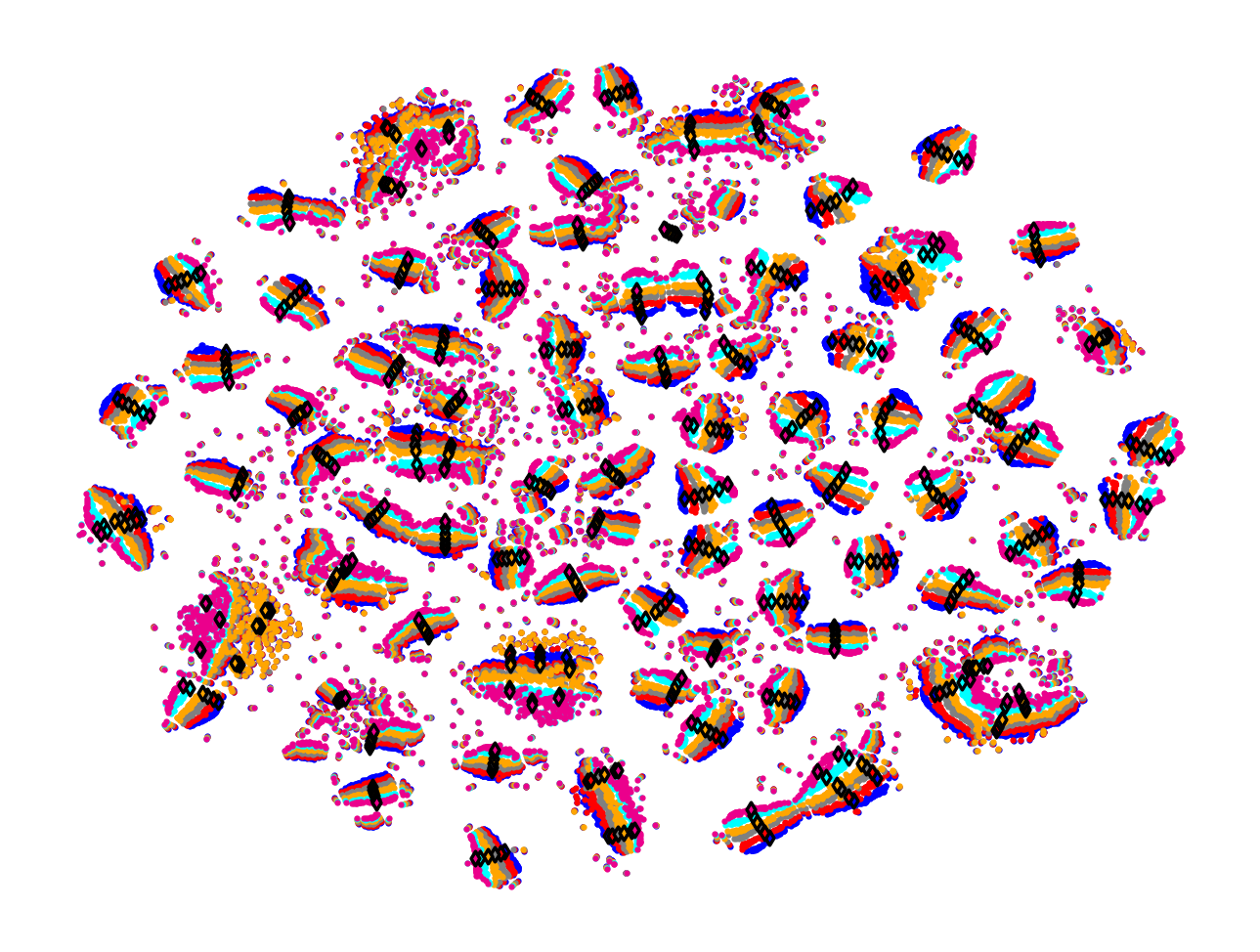}
         \caption{60th epoch, $\# \:of\: classes=6C$}
         \label{fig:vis_e}
     \end{subfigure}
     \begin{subfigure}[b]{0.32\linewidth}
         \centering
         \includegraphics[width=1.0\columnwidth]{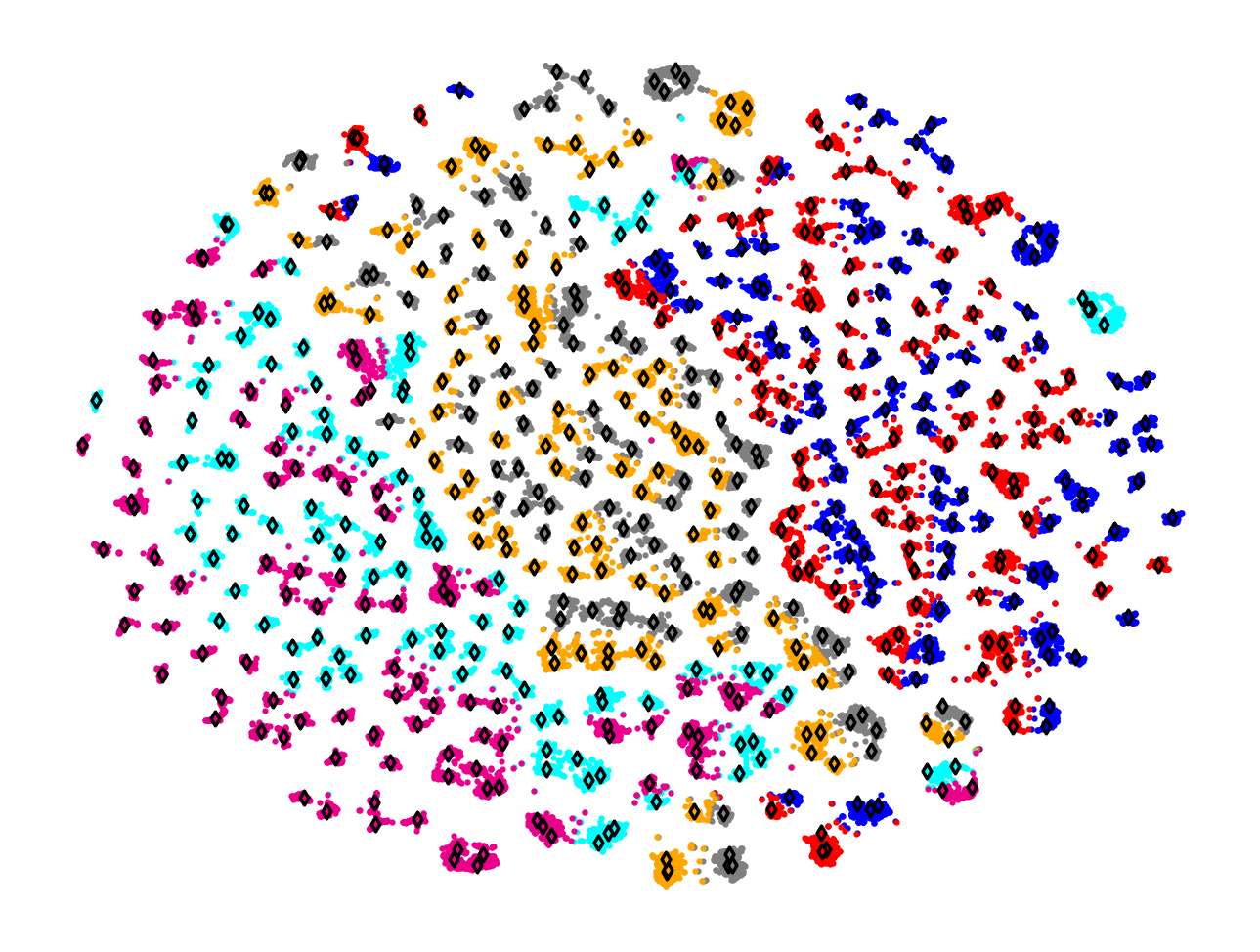}
         \caption{200th epoch, $\# \:of\: classes=6C$}
         \label{fig:vis_f}
     \end{subfigure}
\vspace{-2mm}
\caption{t-SNE visualization of 512-dimensional embedding space. Embedding features are extracted by model trained with MemVir(5,100) on CARS196 training data. Each color indicates step for embedding features.}
\label{fig:tsne}
\end{figure*}

In comparison, the gradient of MemVir + softmax loss over the embedding feature $x_i$ can be inducted as follows:
\begin{eqnarray}
\frac{\partial l_{MemVir}(x_i, y_i)}{\partial x_i} & = & W_{y_i} - \frac{\sum_{j=1}^{(N+1)C} e^{\alpha(x_i, j)} W_j}{\sum_{j=1}^{(N+1)C} e^{\alpha(x_i, j)}} \nonumber \\
        & \approx & W_{y_i} - \frac{\sum_{n=0}^N e^{\alpha(x_i, y_i^{(n)})} W_{y_i^{(n)}}}{\sum_{j=1}^{(N+1)C} e^{\alpha(x_i, j)}} \nonumber \\
        & = & \tau_0 W_{y_i} + \sum_{n=1}^{N} \tau_{n} W_{y_i^{(n)}},
\end{eqnarray}

\begin{eqnarray}
\tau_{0} = 1 - \frac{e^{\alpha(x_i, y_i)}}{\sum_{j=1}^{(N+1)C} e^{\alpha(x_i, j)}}, 
\tau_{n} = - \frac{e^{\alpha(x_i, y_i^{(n)})}}{\sum_{j=1}^{(N+1)C} e^{\alpha(x_i, j)}}
\end{eqnarray}
where, $y_i^{(n)} (n>0)$ are virtual classes and $y_i^{(0)} = y_i$.
It is obvious that $\tau_0 > 0$.
However, $\tau_0$ would not be close to zero whether $x_i$ is nearby $W_{y_i}$ or not, because the denominator of $\tau_0$ would be large as the virtual classes are close to $W_{y_i}$.
As illustrated in Figure~\ref{fig:proof_memvir}, this makes it difficult for $x_i$ to get highly close to $W_{y_i}$ and thus, alleviates the phenomenon of the embedding feature becoming extremely close to the target $W_{y_i}$.
In addition, because $\tau_n < 0$, $x_i$ tries to get farther away from the virtual classes $W_{y_i^{(n)}}$.
Thus, the alleviation would be more extensive and can effectively ease the intense focus of the softmax loss, leading to a more substantial generalization.
This is empirically shown in Figure~\ref{fig:generalization}.
The baseline gradually increases the similarity between the embeddings and corresponding class weights.
By contrast, when MemVir is turned on at step $U$, the similarity is slightly degraded by alleviating the strong focus on the seen classes, and better generalization is achieved as shown in Figure~\ref{fig:gen_acc}.
The detailed induction is provided in the supplementary Section B.1.

\begin{figure*}[t!h!]
     \centering
     \begin{subfigure}[b]{0.3\linewidth}
         \centering
         \includegraphics[width=1.0\columnwidth]{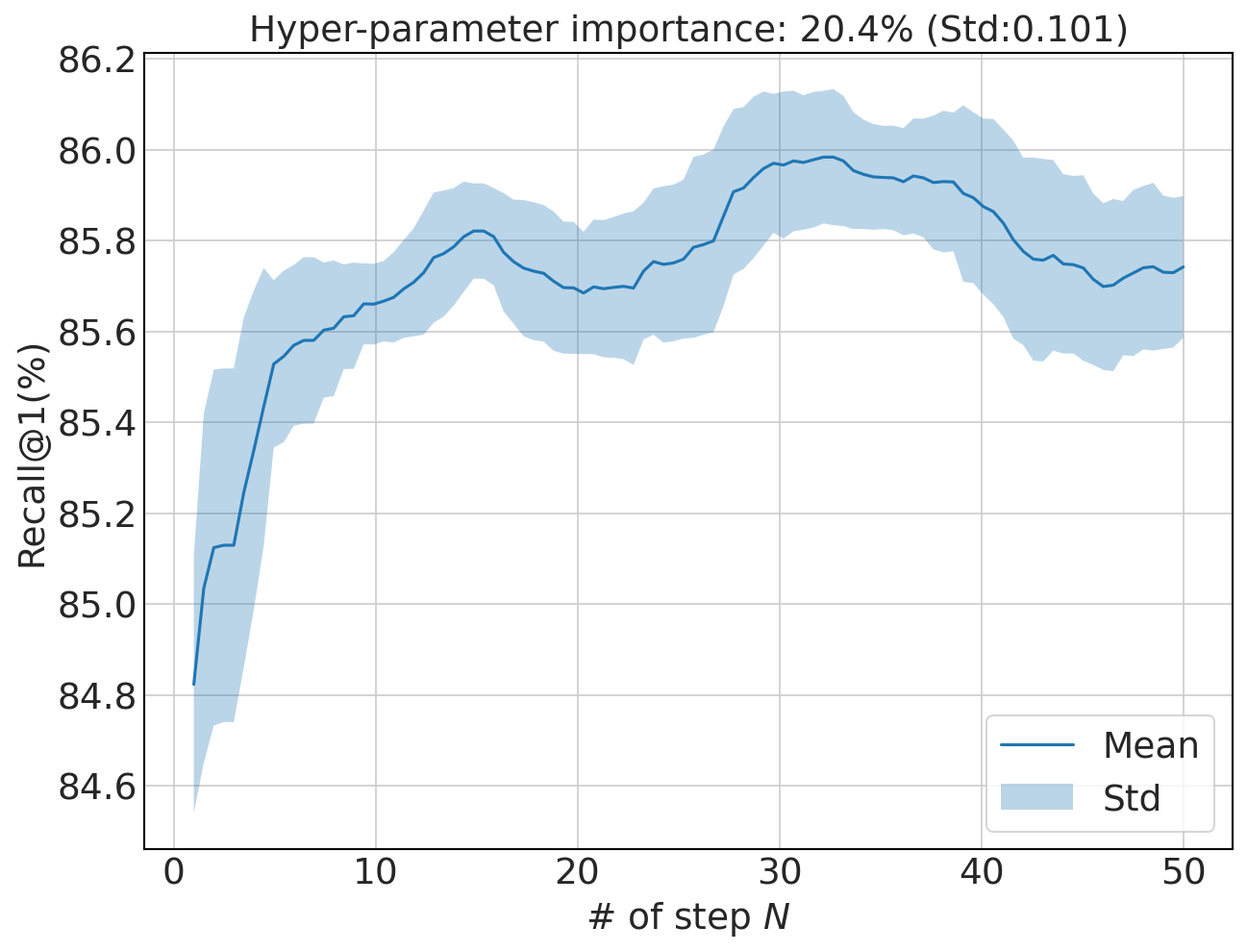}
         \caption{Impact of number of step $N$}
         \label{fig:param_step}
     \end{subfigure}
     \begin{subfigure}[b]{0.3\linewidth}
         \centering
         \includegraphics[width=1.0\columnwidth]{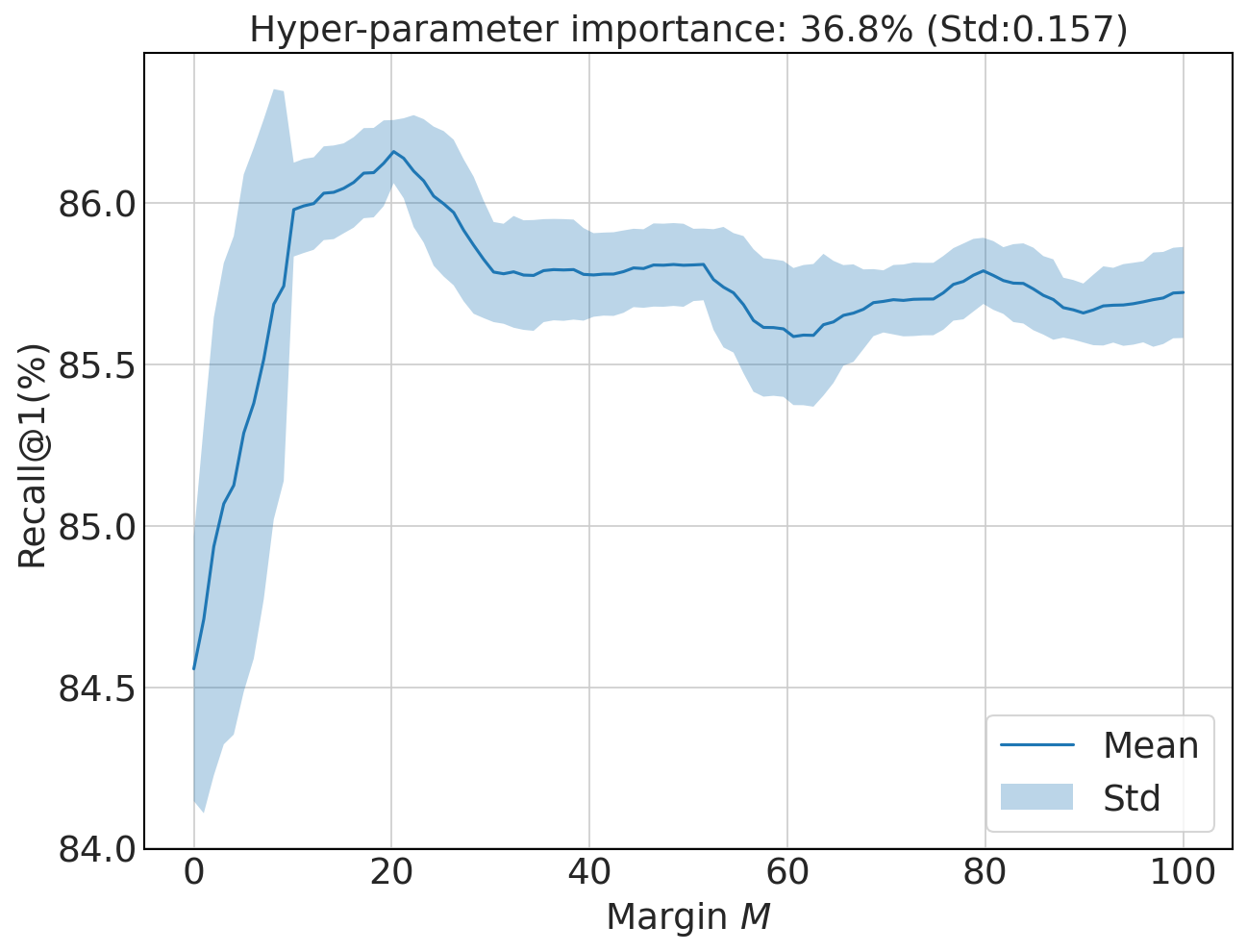}
         \caption{Impact of margin $M$}
         \label{fig:param_margin}
     \end{subfigure}
     \begin{subfigure}[b]{0.3\linewidth}
         \centering
         \includegraphics[width=1.0\columnwidth]{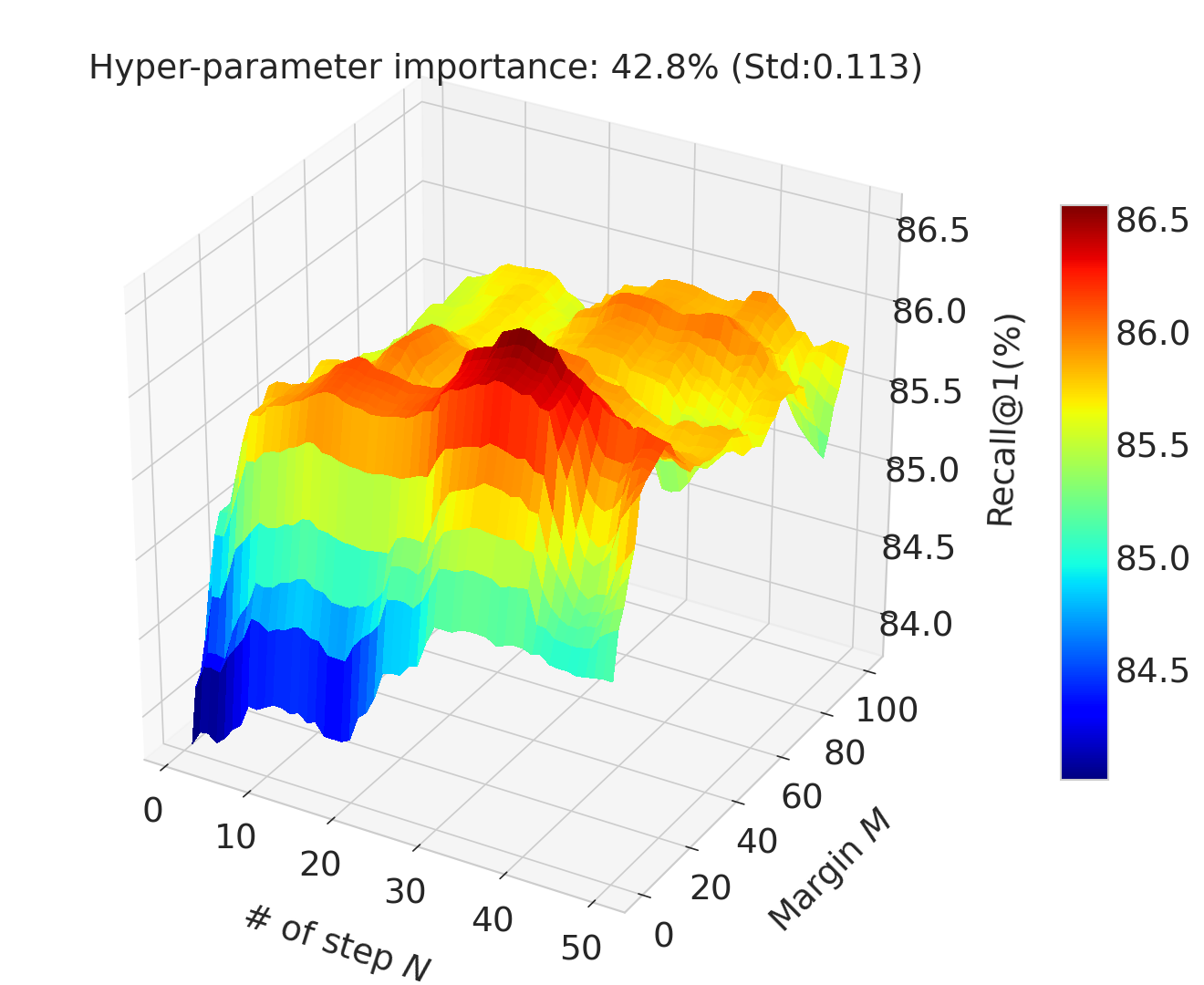}
         \caption{Impact of pair-wise interaction}
         \label{fig:param_pair}
     \end{subfigure}
\caption{We use fANOVA~\cite{hutter2014efficient} to estimate the impact of hyper-parameters. Reported performances are predicted values from random forest of fANOVA, which is trained with experimental results of MemVir on CARS196.}
\label{fig:param}
\end{figure*}

\section{Experiments}
In this section, we conduct a series of experiments to analyze and validate the effectiveness of MemVir.
Please refer to the supplementary material for additional experiments: analysis of memory and computational cost (Section D.1), impact of learning rate (Section D.2), impact of warm-up (Section D.3), robustness to input deformation (Section D.4), impact of embeddings and class weights in virtual class (Section D.6), and more.

\subsection{Experimental Setting}
We use three popular datasets for evaluation in DML: CUB-200-2011 (CUB200)~\cite{wah2011caltech}, CARS196~\cite{krause20133d}, and Standford Online Products (SOP)~\cite{oh2016deep}.
We perform two types of evaluation procedures: \textit{conventional evaluation} and \textit{MLRC evaluation}.
Conventional evaluation is based on the common training and evaluation procedure described in~\cite{oh2016deep,kim2020proxy}.
All experiments are conducted on an Inception network with batch normalization~\cite{ioffe2015batch} and a 512-dimensional embedding feature.
A batch size of 128, the Adam optimizer~\cite{kingma2014adam} with a learning rate of $10^{-4}$, and warm-up epoch $U_e=50$ are adopted unless otherwise noted in the experiment.
Considering recent works~\cite{musgrave2020metric,fehervari2019unbiased} that have proposed improved evaluation procedures for fairness, we include the MLRC evaluation protocol~\cite{musgrave2020metric}.
In MLRC evaluation, the procedure includes hyper-parameter search with 4-fold cross-validation, ensemble evaluation, and the usage of fair metrics (P@1, RP, and MAP@R).
Please refer to supplementary Section C for details regarding the datasets and implementation.

\subsection{Embedding Space Visualization}
In Figure~\ref{fig:tsne}, we visualize the embedding space of the training data via t-SNE~\cite{maaten2008visualizing} to present how MemVir learns the embedding space.
At the 50th epoch in Figure~\ref{fig:vis_b}, the model has been trained with only actual classes and obtains sparse embedding space with concentration on the actual classes.
When all virtual classes are added at the 60th epoch in Figure~\ref{fig:vis_e}, virtual classes tend to be close to the actual classes and the embedding space is still sparse as in Figure~\ref{fig:vis_b}.
This demonstrates that the model is not fully utilizing the embedding space and is highly focused on the seen classes.
After enough epochs of training, at the 200th epoch in Figure~\ref{fig:vis_f}, the model obtains dense embedding space with sufficient discriminative power over all actual and virtual classes.
To sum up, MemVir offers better utilization of embedding space by alleviating strong focus on seen classes for generalization.
We include extended visualization in supplementary Section D.8.

\begin{table}[t!]
\centering
\begin{subtable}[h]{1.0\columnwidth}
\begin{adjustbox}{width=1.0\columnwidth,center}
\begin{tabular}{lcccccccccc}
\toprule
Batch size      & 8   & 16   & 32   & 64   & 128   & 256 & 512   & 1024  \\
\midrule
Norm-softmax  & 79.1 & 82.8 & 83.1 & \underline{\textbf{83.5}} & 83.3 & 82.8 & 81.0 & 78.5 \\
+ \textit{MemVir} & 80.4  & \underline{83.6} & 85.0 & \textbf{85.5} & 85.0 & 85.0 & 84.8 & 84.6 \\
Diff          &  \textcolor{Green}{+1.3} &  \textcolor{Green}{+0.8}  & \textcolor{Green}{+1.9} & \textcolor{Green}{+2.0} & \textcolor{Green}{+1.7} & \textcolor{Green}{+2.2} & \textcolor{Green}{+3.8} & \textbf{\textcolor{Green}{+6.1}} \\ 
\bottomrule
\end{tabular}
\end{adjustbox}
\caption{Impact of batch size.}
\label{table:batch}
\end{subtable}
\hfill
\centering
\begin{subtable}[h]{1.0\columnwidth}
\begin{adjustbox}{width=1.0\columnwidth,center}
\begin{tabular}{lcccccccccc}
\toprule
Class ratio (\%) & 10   & 20   & 30   & 40   & 50   & 60   & 70   & 80   & 90   & 100  \\
\midrule
Norm-softmax    & 56.4 & 67.3 & 69.6 & 74.8 & 77.7 & 78.8 & 79.4 & 81.7 & 82.0 & \underline{\textbf{83.3}} \\
+ \textit{MemVir}   & 58.5 & 70.1 & 72.8 & 77.2 & 80.0 & 81.3 & 82.6 & \underline{83.8} & 84.1 & \textbf{85.0} \\
Diff            & \textcolor{Green}{+2.1} & \textcolor{Green}{+2.8} & \textbf{\textcolor{Green}{+3.2}} & \textcolor{Green}{+2.4} & \textcolor{Green}{+2.3} & \textcolor{Green}{+2.5} & \textbf{\textcolor{Green}{+3.2}} & \textcolor{Green}{+2.1} & \textcolor{Green}{+2.1} & \textcolor{Green}{+1.7} \\
\bottomrule
\end{tabular}
\end{adjustbox}
\caption{Impact of number of classes.}
\label{table:class}
\end{subtable}
\caption{Impact of batch size and number of classes on CARS196 dataset. We report Recall@1(\%) performance and underline when MemVir(1,100) exceeds the best performance of the baseline Norm-softmax.}
\label{table:class_batch}
\end{table}

\subsection{Impact of Batch Size and Number of Classes}
\label{sec:cls_batch}
One advantage of MemVir is that it can utilize augmented information, including an increased number of embedding features and classes without additional feature extraction.
To see the impact of the number of embedding features and classes, we conduct experiments by varying the batch size and number of classes, where the training classes are randomly sampled by class ratio.
As shown in Table~\ref{table:batch}, the performance of the Norm-softmax baseline increases from the batch size of 8 to 64 and then decreases after, indicating that the increase in the batch size does not guarantee performance improvement~\cite{krizhevsky2014one, goyal2017accurate}.
Applying MemVir to the baselines allows the models to learn with twice the number of embedding features by the virtual classes.
MemVir yields performance improvement by $2.5\%$ on average and exceeds the best performance of the baseline of batch size 64 with the batch size of only 16.
Moreover, we observe that using MemVir is more robust to performance degradation due to the large batch size.
As shown in Table~\ref{table:class}, decreasing the class ratio degrades the performance of the Norm-softmax baseline from 83.3\% to 56.4\%.
With MemVir, which doubles the number of classes with virtual classes, we observe that the performance increases by an average of 2.4\% and exceeds the best performance of the baseline with only 80\% of the classes.

\begin{table*}[t!]
\begin{adjustbox}{width=1.0\textwidth,center}
\begin{tabular}{>{\rowmac}l>{\rowmac}c>{\rowmac}c>{\rowmac}c>{\rowmac}c>{\rowmac}c>{\rowmac}c>{\rowmac}c>{\rowmac}c>{\rowmac}c<{\clearrow}}
\toprule
                  & \multicolumn{3}{c}{CUB200}                 & \multicolumn{3}{c}{CARS196}                & \multicolumn{3}{c}{SOP}                    \\ \cline{2-10} 
Method              & P@1          & RP           & MAP@R        & P@1          & RP           & MAP@R        & P@1          & RP           & MAP@R        \\
\midrule
Norm-softmax~\cite{wang2017normface}      & 65.65 $\pm$ 0.30 & 35.99 $\pm$ 0.15 & 25.25 $\pm$ 0.13 & 83.16 $\pm$ 0.25 & 36.20 $\pm$ 0.26 & 26.00 $\pm$ 0.30 & 75.67 $\pm$ 0.17 & 50.01 $\pm$ 0.22 & 47.13 $\pm$ 0.22 \\
 \textit{MemVir} + Norm-softmax \setrow{\bfseries} & 69.22 $\pm$ 0.15 & 37.92 $\pm$ 0.16 & 27.10 $\pm$ 0.13 & 85.81 $\pm$ 0.18 & 38.78 $\pm$ 0.19 & 28.92 $\pm$ 0.17 & 75.77 $\pm$ 0.20 & 50.24 $\pm$ 0.22 & 47.45 $\pm$ 0.25 \\
CosFace~\cite{wang2018cosface}           & 67.32 $\pm$ 0.32 & 37.49 $\pm$ 0.21 & 26.70 $\pm$ 0.23 & 85.52 $\pm$ 0.24 & 37.32 $\pm$ 0.28 & 27.57 $\pm$ 0.30 & 75.79 $\pm$ 0.14 & 49.77 $\pm$ 0.19 & 46.92 $\pm$ 0.19 \\
 \textit{MemVir} + CosFace  \setrow{\bfseries}    & 69.79 $\pm$ 0.26 & 37.85 $\pm$ 0.23 & 27.08 $\pm$ 0.28 & 87.57 $\pm$ 0.13 & 39.10 $\pm$ 0.21 & 29.56 $\pm$ 0.26 & 75.88 $\pm$ 0.27 & 49.95 $\pm$ 0.37 & 47.18 $\pm$ 0.38 \\
ArcFace~\cite{deng2019arcface}           & 67.50 $\pm$ 0.25 & 37.31 $\pm$ 0.21 & 26.45 $\pm$ 0.20 & 85.44 $\pm$ 0.28 & 37.02 $\pm$ 0.29 & 27.22 $\pm$ 0.30 & \textbf{76.20} $\pm$ \textbf{0.27} & 50.27 $\pm$ 0.38 & 47.41 $\pm$ 0.40 \\
 \textit{MemVir} + ArcFace                 & \textbf{69.33} $\pm$ \textbf{0.41} & \textbf{37.82} $\pm$ \textbf{0.28} & \textbf{26.96} $\pm$ \textbf{0.25} & \textbf{88.02} $\pm$ \textbf{0.18} & \textbf{39.12} $\pm$ \textbf{0.15} & \textbf{29.63} $\pm$ \textbf{0.15} & 76.05 $\pm$ 0.30 & \textbf{50.56} $\pm$ \textbf{0.33} & \textbf{47.75} $\pm$ \textbf{0.32} \\
Proxy-NCA~\cite{movshovitz2017no}         & 65.69 $\pm$ 0.43 & 35.14 $\pm$ 0.26 & 24.21 $\pm$ 0.27 & 83.56 $\pm$ 0.27 & 35.62 $\pm$ 0.28 & 25.38 $\pm$ 0.31 & 75.89 $\pm$ 0.17 & 50.10 $\pm$ 0.22 & 47.22 $\pm$ 0.21 \\
 \textit{MemVir} + Proxy-NCA  \setrow{\bfseries}  & 69.25 $\pm$ 0.32 & 37.31 $\pm$ 0.12 & 26.43 $\pm$ 0.17 & 87.02 $\pm$ 0.15 & 38.51 $\pm$ 0.15 & 28.76 $\pm$ 0.16 & 76.97 $\pm$ 0.31 & 50.81 $\pm$ 0.26 & 48.02 $\pm$ 0.27 \\
Proxy-anchor~\cite{kim2020proxy}      & 69.73 $\pm$ 0.31 & 38.23 $\pm$ 0.37 & 27.44 $\pm$ 0.35 & 86.20 $\pm$ 0.21 & 39.08 $\pm$ 0.31 & 29.37 $\pm$ 0.29 & 75.37 $\pm$ 0.15 & 50.19 $\pm$ 0.14 & 47.25 $\pm$ 0.15 \\
 \textit{MemVir} + Proxy-anchor \setrow{\bfseries} & 69.81 $\pm$ 0.28 & 38.57 $\pm$ 0.14 & 27.83 $\pm$ 0.16 & 86.40 $\pm$ 0.18 & 40.27 $\pm$ 0.20 & 30.58 $\pm$ 0.20 & 77.80 $\pm$ 0.17 & 53.21 $\pm$ 0.12 & 50.35 $\pm$ 0.13 \\
\bottomrule
\end{tabular}
\end{adjustbox}
\vspace{-0.5em}
\caption{\textbf{[MLRC evaluation]} Performance (\%) on three famous datasets in image retrieval task. We report the performance of concatenated 512-dim over 10 training runs. Bold numbers indicate the best score within the same loss and dataset.}
\label{table:mlrc}
\end{table*}

\begin{table}[t!]
\centering
\begin{adjustbox}{width=1.0\columnwidth,center}
\begin{tabular}{lcccccc}
\toprule
\multicolumn{1}{c}{}   & \multicolumn{3}{c}{CARS196} & \multicolumn{3}{c}{SOP} \\ \cline{2-7} 
Method   & R@1     & R@2     & R@4     & R@1    & R@10  & R@100  \\
\midrule
SoftMax                & 78.3    & 86.4    & 91.9    & 76.6   & 89.4  & 95.8   \\
Virtual SoftMax        & 75.1    & 84.1    & 90.1    & 74.5   & 87.9  & 94.8   \\
\textit{MemVir} + SoftMax        & \textbf{79.2}    & \textbf{87.0}    & \textbf{92.1}    & \textbf{78.9}   & \textbf{90.6}  & \textbf{96.2}   \\
\midrule
ArcFace                & 78.8    & 86.4    & 91.7    & 76.9   & 89.1  & 95.0   \\
XBM + ArcFace          & 78.9    & 86.2    & 91.9    & 78.1   & 89.7  & 95.8   \\
BroadFace + ArcFace    & 79.5    & 87.3    & 92.0    & 80.2   & 91.0  & 95.9   \\
\textit{MemVir} + ArcFace        & \textbf{80.7}    & \textbf{88.1}    & \textbf{92.7}    & \textbf{80.8}   & \textbf{91.3}  & \textbf{96.5}   \\
\midrule
CurricularFace         & 79.9    & 87.3    & 92.0    & 79.8   &  90.7   &  95.6  \\
\textit{MemVir} + CurricularFace & \textbf{81.0}    &  \textbf{87.9} &  \textbf{92.9}     & \textbf{81.3}   & \textbf{91.7}  & \textbf{98.8}  \\
\bottomrule
\end{tabular}
\end{adjustbox}
\vspace{-0.5em}
\caption{Performance (\%) comparison with related methods on CARS196 and SOP dataset.}
\label{table:related}
\end{table}

\subsection{Impact of Hyper-parameters}
\label{sec:param}

For hyper-parameter analysis, we use the fANOVA framework~\cite{hutter2014efficient}, which can estimate the pattern and importance of each hyper-parameter and pair-wise interaction.
We report the hyper-parameter analysis of CUB200 and SOP as well as the details of the fANOVA in the supplementary Section C.3 and D.5.
As illustrated in Figure~\ref{fig:param}, the performance on CARS196 improves as the number of steps $N$ increases.
The performance improves until the margin $M=20$, and then stabilizes after a slight degradation.
However, the patterns of the impact of the hyper-parameters differ for each dataset because the characteristics of each dataset and the number of classes are diverse. 
We observe two common patterns among all datasets.
First, a margin $M$ larger than zero is typically better than $M=0$; this is because classes from adjacent steps would be too similar to act as different classes and hence become distractions.
Second, $N$ exceeding one is typically better than $N=1$.
This is because by using more steps $N$, the effect of CL can be exploited more effectively by scheduling addition of virtual classes with a longer time.

\subsection{Comparison with Related Methods}

We compare MemVir with related methods from image recognition task, including the virtual class (Virtual softmax~\cite{chen2018virtual}), the memory-based (BroadFace~\cite{kim2020broadface}), and the CL (CurricularFace~\cite{huang2020curricularface}) methods.
Also, we include XBM~\cite{wang2020cross} from DML to compare with BroadFace.
For a fair comparison, we follow the experimental setting of~\cite{kim2020broadface,huang2020curricularface}, which consists of a stochastic gradient descent (SGD) optimizer, a learning rate of $5 \times 10^{-3}$, a batch size of 512, and the ResNet50 backbone~\cite{he2016deep}.
As presented in Table~\ref{table:related}, Virtual softmax degrades the performance, whereas MemVir + softmax improves the performances of both datasets.
When we combine XBM with ArcFace, we observe performance degradation when the memory size is large, as reported in BroadFace~\cite{kim2020broadface}.
The performance can be further improved by adding compensation technique and gradient control presented in BroadFace.
However, exploiting the memorized features as virtual classes in MemVir shows a higher performance boost than just utilizing them for the increased number of instances in XBM and BroadFace.
Considering that CurricularFace has already included the idea of CL, MemVir can improve the performance even further by providing virtual classes as harder cases.
Moreover, it is noteworthy that the experimental results show the flexibility of MemVir for different types of backbones and optimizers.
Extended experiments with different experimental settings are presented in the supplementary Section D.7.

\begin{table}[t]
\begin{adjustbox}{width=1.0\columnwidth,center}
\begin{tabular}{>{\rowmac}l>{\rowmac}c>{\rowmac}c>{\rowmac}c<{\clearrow}}
\toprule
Method     & CUB200  & CARS196 & SOP\\
\midrule
Multi-similarity (MS)$^{\dagger}$~\cite{wang2019multi}   & 64.5        & 82.1        & 76.3        \\
SoftTriple~\cite{qian2019softtriple}      & 65.4        & 84.5        & 78.3        \\
ProxyGML~\cite{zhu2020fewer}         & 66.6        & 85.5        & 78.0        \\
Symm~\cite{gu2020symmetrical} + MS~\cite{sohn2016improved}  & 64.9        & 82.4        & 76.9        \\
EE~\cite{ko2020embedding} + MS~\cite{wang2019multi}   & 65.1        & 82.9        & 77.0        \\
XBM~\cite{wang2020cross} + Contrastive~\cite{hadsell2006dimensionality}   & 65.8        & 82.0        & 79.5        \\
\midrule
Softmax           & 64.2        & 81.5        & 76.3        \\
\textit{MemVir} + Softmax     & 66.8 \textcolor{Green}{(+2.6)} & 86.5 \textcolor{Green}{(+5.0)} & 77.8 \textcolor{Green}{(+1.5)} \\
Norm-softmax~\cite{wang2017normface}      & 64.9        & 83.3        & 78.6        \\
\textit{MemVir} + Norm-softmax   & 67.3 \textcolor{Green}{(+2.4)} & \textbf{86.8 \textcolor{Green}{(+3.5)}} & 79.6 \textcolor{Green}{(+1.0)} \\
CosFace~\cite{wang2018cosface}           & 65.7        & 83.6        & 78.6        \\
\textit{MemVir} + CosFace     & 67.7 \textcolor{Green}{(+2.0)} & 86.6 \textcolor{Green}{(+3.0)} & 79.7 \textcolor{Green}{(+1.1)} \\
ArcFace~\cite{deng2019arcface}        & 66.1        & 83.7        & 78.8        \\
 \textit{MemVir} + ArcFace     & 67.4 \textcolor{Green}{(+1.3)} & 86.5 \textcolor{Green}{(+2.8)} & \textbf{80.0 \textcolor{Green}{(+1.2)}} \\
Proxy-NCA~\cite{movshovitz2017no}        & 64.3        & 82.0        & 78.1        \\
 \textit{MemVir} + Proxy-NCA   & 68.3 \textcolor{Green}{(+4.0)} & 86.5 \textcolor{Green}{(+4.5)} & 79.2 \textcolor{Green}{(+1.1)} \\
Proxy-anchor$^{\dagger}$~\cite{kim2020proxy}     & 67.7        & 84.9        & 78.9        \\
 \textit{MemVir} + Proxy-anchor  & \textbf{69.0 \textcolor{Green}{(+1.3)}} & 86.7 \textcolor{Green}{(+1.8)} & 79.7 \textcolor{Green}{(+0.8)} \\
\bottomrule
\end{tabular}
\end{adjustbox}
\vspace{-0.5em}
\caption{\textbf{[Conventional evaluation]} Recall@1 (\%) on three famous datasets in image retrieval task. $^{\dagger}$ denotes evaluation in a fair setting described in supplementary Section C.2.1.\vspace{-0.5em}}
\label{table:sota}
\end{table}

\subsection{Comparison with State-of-the-art}
Finally, we compare the proposed method with state-of-the-art methods in DML.
In the conventional evaluation shown in Table~\ref{table:sota}, every softmax variant and proxy-based loss combined with MemVir show significantly improved performance in every dataset.
The average performance improvements are $2.3\%$, $3.4\%$, and $1.1\%$ for CUB200, CARS196 and SOP, respectively.
In comparison with the memory-based (XBM), sample generation (Symm, EE), and other recent methods (MS, SoftTriple and ProxyGML), MemVir shows competitive performance in all datasets.
Even in the MLRC evaluation shown in Table~\ref{table:mlrc}, which is specifically designed in terms of fairness, MemVir improves performance in every dataset and metric substantially.
These results demonstrate the flexibility and effectiveness of MemVir in DML.
Please refer to the supplementary Section D.9 for extended results of the metrics and comparisons with existing methods in conventional evaluation, as well as additional performance report of separated 128-dim in MLRC evaluation.

\section{Conclusion}
In this paper, we have presented a novel training strategy that exploits memory-based virtual classes and incorporates the idea of CL.
Theoretical and empirical analysis demonstrates that employing virtual classes as augmented information help achieve better generalization by alleviating a strong focus on seen classes.
Furthermore, we show that gradually increasing the learning difficulty by slowly adding virtual classes improves the training process and final performance.
Considering that MemVir is easily applicable to existing loss functions for better generalization, it is hence a competitive training strategy in DML.

\paragraph{Acknowledgement.}
We would like to thank all members of Visual Search team at NAVER/LINE for helpful comments and feedback.
We are grateful to Yoonjae Cho, Tae Kwan Lee, and Jingeun Lee for the detailed review of the paper.

{\small
\bibliographystyle{ieee_fullname}
\bibliography{egbib}
}

\clearpage



\section*{Supplementary material}

# Learning with Memory-based Virtual Classes for Deep Metric Learning

*Supplementary Material*

github.com/navervision/MemVir

# Contents

## A. Loss Functions

In this section, we briefly describe softmax variant and proxy-based losses used in our study, as well as the hyper-parameters of each loss. For notation, we refer to the embedding features as $x_i$ and corresponding label as $y_i$. Each loss function $l(\cdot)$ is written based on the generalized from of objective function:

$$\mathcal{L}(X, W) = -\frac{1}{|X|} \sum_{i=1}^{|X|} l(x_i, y_i), \tag{i}$$

where $X$ and $W$ are sets of embedding features and class weights, respectively.

**Proxy-NCA [19]** Typically, pair-based losses suffer from sampling issues such that sampling tuples heavily affects the training convergence. To address this problem, Proxy-NCA loss introduces class proxies, which represent each class. In this way, we can sample only one anchor and compare it against the corresponding positive and negative class proxies. It is noteworthy that class proxies have the same meaning as class weights from the softmax variants theoretically and practically; thus, we use the term 'class weights' to include class representatives or proxies in this paper. The Proxy-NCA loss can be formulated as:

$$l_{proxy-NCA}(x_i, y_i) = \log \frac{e^{-d(x_i, W_{y_i})}}{\sum_{j=1}^{C} e^{-d(x_i, W_j)}}, \tag{ii}$$

where $d(a, b) = \| a - b \|_2$ is the Euclidean distance between $a$ and $b$.

**Proxy-anchor [11]** Unlike typical softmax variants and proxy-based losses, Proxy-anchor loss uses each proxy as an anchor and considers its relations with all samples in a batch. We define $X_w^+$ and $X_w^-$ as the set of positive and negative embedding features of each proxy (class weight) $w$, respectively, and $W^+$ as the set of positive proxies of data in the mini-batch. Because of its peculiar structure, we formulate the Proxy-anchor loss based on the mini-batch as follows:

$$\mathcal{L}_{proxy-anchor}(X, W) = \frac{1}{|W^+|} \sum_{w \in W^+} \log \left\{ 1 + \sum_{x \in X_w^+} e^{-\gamma(s(x,w)-\delta)} \right\} + \frac{1}{|W|} \sum_{w \in W} \log \left\{ 1 + \sum_{x \in X_w^-} e^{\gamma(s(x,w)+\delta)} \right\}, \tag{iii}$$

where $s(a, b) = a^T b$ denotes the cosine similarity between $a$ and $b$, $\gamma$ is a scaling factor, and $\delta$ is a margin parameter. We use $\gamma = 46$ and $m = 0.1$ for our hyper-parameters.

**CosFace [30]** CosFace loss reformulates the softmax loss as a cosine loss by $l_2$ normalizing the embedding features and class weights, which is equivalent to the Norm-softmax loss, and defines a decision margin in cosine space as:

$$l_{cosface}(x_i, y_i) = \log \frac{e^{\gamma(\cos(\theta_{y_i})-m)}}{e^{\gamma(\cos(\theta_{y_i})-m)} + \sum_{j=1, j \neq y_i}^{C} e^{\gamma \cos \theta_j}}, \tag{iv}$$

where $\gamma$ is a scale and $m$ is a margin parameter. For our hyper-parameters, we use $\gamma = 28$ and $m = 0.1$.

**ArcFace [2]** Similar to CosFace loss, ArcFace loss transforms the Norm-softmax loss function by applying an angular margin between the embedding $x_i$ and the corresponding class weight $W_{y_i}$ for each class. ArcFace loss can be formulated as:

$$l_{arcface}(x_i, y_i) = \log \frac{e^{\gamma \cos(\theta_{y_i}+m)}}{e^{\gamma \cos(\theta_{y_i}+m)} + \sum_{j=1, j \neq y_i}^{C} e^{\gamma \cos \theta_j}}, \tag{v}$$

where we use the scale $\gamma = 24$ and the margin $m = 0.1$.

**CurricularFace [7]** CurricularFace incorporates the idea of curriculum learning into the ArcFace loss to adjust the relative importance of easy and hard samples during training. The loss function of CurricularFace is formulated as follows:

$$l_{curricularface}(x_i, y_i) = \log \frac{e^{\gamma T(\cos\theta_{y_i})}}{e^{\gamma T(\cos\theta_{y_i})} + \sum_{j=1, j \neq y_i}^{C} e^{\gamma N(t, \cos\theta_j)}}, \quad \text{(vi)}$$

$$T(cos\theta_{y_i}) = cos(\theta_{y_i} + m), \quad \text{(vii)}$$

$$N(t, cos\theta_{y_j}) = \begin{cases} cos\theta_{y_j}, & T(cos\theta_{y_i}) - cos\theta_j \geq 0 \\ cos\theta_{y_j}(t^{(k)} + cos\theta_j), & T(cos\theta_{y_i}) - cos\theta_j < 0, \end{cases} \quad \text{(viii)}$$

where $T(\cdot)$ and $N(\cdot)$ are modulation functions for the positive and negative cosine similarities, respectively. The parameter $t^{(k)}$ of the $k$-th step is computed as follows:

$$t^{(k)} = \alpha r^{(k)} + (1 - \alpha) t^{(k)}, \quad \text{(ix)}$$

where $r^{(k)}$ is the average of the positive cosine similarities and $\alpha$ is the momentum parameter. For our hyper-parameters, we use $\alpha = 0.99$, $\gamma = 26$ and $m = 0.3$.

## B. Details of MemVir

### B.1. Proof of Gradient Analysis for Generalization

We provide a detailed proof of gradient analysis for generalization, which is discussed in Section 3.3.3. The proof is shown by comparing the gradient descent of softmax and MemVir.

**Gradient Descent of Softmax:** As mentioned in Equation 6 of the main paper, the softmax loss can be written as follows:

$$l_{softmax}(x_i, y_i) = \log \frac{e^{\alpha(x_i, y_i)}}{\sum_{j=1}^{C} e^{\alpha(x_i, j)}}, \quad \text{(x)}$$

where $\alpha(x_i, j) = W_j^T x_i$. The gradient of the softmax loss over the embedding feature $x_i$ can be inducted as follows:

$$\frac{\partial l_{softmax}(x_i, y_i)}{\partial x_i} = \frac{\sum_{j=1}^{C} e^{\alpha(x_i,j)}}{e^{\alpha(x_i,y_i)}} \frac{W_{y_i} e^{\alpha(x_i,y_i)} \sum_{j=1}^{C} e^{\alpha(x_i,j)} - e^{\alpha(x_i,y_i)} \sum_{j=1}^{C} W_j e^{\alpha(x_i,j)}}{(\sum_{j=1}^{C} e^{\alpha(x_i,j)})^2}$$

$$= W_{y_i} - \frac{\sum_{j=1}^{C} e^{\alpha(x_i,j)} W_j}{\sum_{j=1}^{C} e^{\alpha(x_i,j)}}. \quad \text{(xi)}$$

For a moderately-trained model, $x_i$ should much more closer to $W_{y_i}$ compared to $W_j$ for $j \neq y_i$, which leads to $\alpha(y_i, i) > \alpha(j, i)$ (*Condition 1*). Besides, for a model with norm-softmax loss, $\alpha(j, i) \in (-\gamma, +\gamma)$, where $\gamma \geq 10$ in our experiment setting (*Condition 2*). Considering *Condition 1* with *Condition 2* together, we can derive the conclusion that $e^{\alpha(x_i, y_i)} >> e^{\alpha(x_i, j)}$ for $j \neq y_i$. By applying such conclusion to the numerator of Equation xi for approximation, following equation can be achieved:

$$\frac{\partial l_{softmax}(x_i, y_i)}{\partial x_i} = W_{y_i} - \frac{\sum_{j=1}^{C} e^{\alpha(x_i,j)} W_j}{\sum_{j=1}^{C} e^{\alpha(x_i,j)}}$$

$$\approx W_{y_i} - \frac{e^{\alpha(x_i, y_i)} W_{y_i}}{\sum_{j=1}^{C} e^{\alpha(x_i,j)}}$$

$$= \tau W_{y_i}, \quad \text{(xii)}$$

$$\tau = 1 - \frac{e^{\alpha(x_i, y_i)}}{\sum_{j=1}^{C} e^{\alpha(x_i,j)}}. \quad \text{(xiii)}$$

It is obvious that $\tau > 0$. Besides, when $x_i \to W_{y_i}$, $e^{\alpha(x_i,y_i)}$ becomes much larger than $e^{\alpha(x_i,j)}$, which leads to $\tau = 1 - \frac{e^{\alpha(x_i,y_i)}}{\sum_{j=1}^{C} e^{\alpha(x_i,j)}} \to 1 - \frac{e^{\alpha(x_i,y_i)}}{e^{\alpha(x_i,y_i)}} = 0$, implying that $x_i$ converges as close to $W_{y_i}$ as possible. This can result in a strong focus on the target weight $W_{y_i}$ and an over-fit to the seen classes of the training data.

**Gradient Descent of MemVir:** For MemVir, the softmax function in Equation x should be re-written as follows:

$$l_{softmax}(x_i, y_i) = \log \frac{e^{\alpha(x_i,y_i)}}{\sum_{j=1}^{(N+1)C} e^{\alpha(x_i,j)}}, \quad \text{(xiv)}$$

where the number of class weights increase from $C$ to $(N+1)C$ compared to softmax because of the existence of the virtual classes. Then, the gradient of MemVir + softmax loss over the embedding feature $x_i$ can be inducted as follows:

$$\begin{aligned}\frac{\partial l_{MemVir}(x_i, y_i)}{\partial x_i} &= \frac{\sum_{j=1}^{(N+1)C} e^{\alpha(x_i,j)}}{e^{\alpha(x_i,y_i)}} \frac{W_{y_i} e^{\alpha(x_i,y_i)} \sum_{j=1}^{(N+1)C} e^{\alpha(x_i,j)} - e^{\alpha(x_i,y_i)} \sum_{j=1}^{(N+1)C} W_j e^{\alpha(x_i,j)}}{(\sum_{j=1}^{(N+1)C} e^{\alpha(x_i,j)})^2} \\ &= W_{y_i} - \frac{\sum_{j=1}^{(N+1)C} e^{\alpha(x_i,j)} W_j}{\sum_{j=1}^{(N+1)C} e^{\alpha(x_i,j)}}\end{aligned} \quad \text{(xv)}$$

During training process, the class weights of past and present tend to be close each other, leading to the conclusion of $e^{\alpha(x_i,y_i)} \gg e^{\alpha(x_i,j)}$ for $j \neq y_i^{(n)}$, where $y_i^{(n)}$ with $n = 1, 2, 3, ..., N$ represents the virtual classes of class $y_i$ from the step $n$ and the term $y_i^{(0)} = y_i$ is used for convenience. By applying such conclusion to the numerator of Equation xv for approximation, following equation can be achieved:

$$\begin{aligned}\frac{\partial l_{MemVir}(x_i, y_i)}{\partial x_i} &= W_{y_i} - \frac{\sum_{j=1}^{(N+1)C} e^{\alpha(x_i,j)} W_j}{\sum_{j=1}^{(N+1)C} e^{\alpha(x_i,j)}} \\ &\approx W_{y_i} - \frac{\sum_{n=0}^{N} e^{\alpha(x_i,y_i^{(n)})} W_{y_i^{(n)}}}{\sum_{j=1}^{(N+1)C} e^{\alpha(x_i,j)}} \\ &= \tau_0 W_{y_i} + \sum_{n=1}^{N} \tau_n W_{y_i^{(n)}},\end{aligned} \quad \text{(xvi)}$$

$$\tau_0 = 1 - \frac{e^{\alpha(x_i,y_i)}}{\sum_{j=1}^{(N+1)C} e^{\alpha(x_i,j)}}, \tau_n = -\frac{e^{\alpha(x_i,y_i^{(n)})}}{\sum_{j=1}^{(N+1)C} e^{\alpha(x_i,j)}} \quad \text{(xvii)}$$

It is obvious that $\tau_0 > 0$. However, $\tau_0$ would not be close to zero whether $x_i$ is nearby $W_{y_i}$ or not, because virtual classes would be close to $W_{y_i}$. This can alleviate the phenomenon of the embedding feature becoming extremely close to the target $W_{y_i}$. In addition, because of $\tau_n$, such alleviation would be more extensive and can effectively ease the intense focus of the softmax loss, leading to a more substantial generalization.

### B.2. Revisiting Slow Drift Phenomena

"*Slow drift*" phenomena have been introduced in [33], which identifies the slow drifting speed of the embeddings by measuring the difference of features for the same instance computed at different training steps. With this observation, [33] suggests using the embeddings of past steps for loss computation of the current step. We reproduce "*slow drift*" phenomena with weight features as follows:

$$\triangle d(W_t, W_{t-m}) := \| W_t - W_{t-(m+1)} \|_2^2, \quad \text{(xviii)}$$

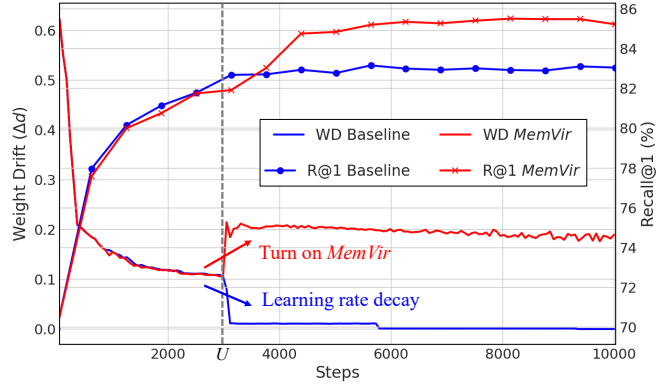

Figure A. Weight drift $\triangle d$ with $m = 100$ and corresponding Recall@1 performance of baseline and MemVir(1,100).

where $W_t$ is weight features of current step $t$ and $W_{t-(m+1)}$ is weight features of past step $t-(m+1)$. As illustrated in Figure A, the weight drift of the baseline is very slow, especially after learning rate decay. We can observe that it converges to local minima with a negligible amount of weight updates after the learning rate decay. On the contrary, when we turn on MemVir instead of learning rate decay, we observe that the weight drift has risen by the enlarged magnitude of gradient; thus, it gives more chance to escape from the local minima for performance improvement.

We further analyze these phenomena by comparing the gradient descent of the softmax loss and MemVir. The softmax loss can be written as follows:

$$
\begin{aligned}
l_{softmax}(x_i, y_i) &= \log \frac{e^{W_{y_i}^T x_i}}{\sum_{j=1}^{C} e^{W_j^T x_i}} \\
&= \log \frac{e^{\alpha(y_i, i)}}{\sum_{j=1}^{C} e^{\alpha(j, i)}},
\end{aligned} \tag{xix}
$$

where $\alpha(j, i) = W_j^T x_i$. Then, the gradient of such loss over $\alpha(j, i)$ can be inducted as follows:

$$
\begin{aligned}
\frac{\partial l_{softmax}(x_i, y_i)}{\partial \alpha(y_i, i)} &= \frac{\sum_{j=1}^{C} e^{\alpha(j, i)} - e^{\alpha(y_i, i)}}{\sum_{j=1}^{C} e^{\alpha(j, i)}} \\
&= 1 - \frac{e^{\alpha(y_i, i)}}{\sum_{j=1}^{C} e^{\alpha(j, i)}}.
\end{aligned} \tag{xx}
$$

Because $e^{\alpha(y_i, i)} >> e^{\alpha(j, i)}$ for $j \neq y_i$, thus, $\frac{\partial l_{softmax}(x_i, y_i)}{\partial \alpha(y_i, i)}$ is close to zero. Considering the weight update is performed by $w = w - \eta \frac{\partial L}{\partial w}$, where $\eta$ is a learning rate, the weight update will be negligible and result in "*slow drift*" phenomena, especially with a small learning rate.

MemVir overcomes such situation by increasing the magnitudes of gradients. The inducted gradient of MemVir over $\alpha(j, i)$ is as follows:

$$
\begin{aligned}
\frac{\partial l_{softmax}(x_i, y_i)}{\partial \alpha(y_i, i)} &= \frac{\sum_{j=1}^{(N+1)C} e^{\alpha(j, i)} - e^{\alpha(y_i, i)}}{\sum_{j=1}^{(N+1)C} e^{\alpha(j, i)}} \\
&= 1 - \frac{e^{\alpha(y_i, i)}}{\sum_{j=1}^{(N+1)C} e^{\alpha(j, i)}} \\
&\approx 1 - \frac{e^{\alpha(y_i, i)}}{e^{\alpha(y_i, i)} + \sum_{n=1}^{N} e^{\alpha(y_i^{(n)}, i)}},
\end{aligned} \tag{xxi}
$$

|  | Softmax | ArcFace | CurricularFace |
| --- | --- | --- | --- |
| CARS196 | (30, 50) | (40,50) | (5,100) |
| SOP | (4, 50) | (10,50) | (9,50) |

(a) Comparison with related methods.

|  | Softmax | Norm-softmax | CosFace | ArcFace | Proxy-NCA | Proxy-anchor |
| --- | --- | --- | --- | --- | --- | --- |
| CUB200 | (5,75) | (10,10) | (25,150) | (40,40) | (10,25) | (5,75) |
| CARS196 | (15,25) | (45,10) | (25,150) | (15,10) | (20,25) | (15,75) |
| SOP | (3,150) | (7,90) | (6,100) | (9,80) | (2,80) | (8,50) |

(b) Conventional evaluation.

|  | Norm-softmax | CosFace | ArcFace | Proxy-NCA | Proxy-anchor |
| --- | --- | --- | --- | --- | --- |
| CUB200 | (15,20) | (35,75) | (40,100) | (40,50) | (15,75) |
| CARS196 | (50,50) | (2,300) | (35,10) | (10,250) | (25,25) |
| SOP | (15,50) | (10,50) | (5,200) | (20,50) | (5,100) |

(c) MLRC evaluation.

Table A. Hyper-parameters ($N$,$M$) of MemVir for each experiment.

where $y_i^{(n)}$ is the index of the weights before $n$ steps. As $e^{\alpha(y_i^{(n)},i)}$ are not close to zero, $\frac{\partial l_{softmax}(x_i,y_i)}{\partial \alpha(y_i,i)}$ will not be close to zero. Thus, the gradient of MemVir will be relatively larger than that of the softmax loss, which gives more chance to escape from the local minima. Note that the magnitudes of gradients can be controlled by the difficulty of curriculum learning, such as hyper-parameters $N$ and $M$.

## C. Details of Experimental Settings

### C.1. Datasets

Throughout the paper, we use three famous benchmarking datasets in deep metric learning (DML) as follows:

- CUB200-2011 (CUB200) [28]: CUB200 contains 11,788 images of birds in 200 classes. We use 5,864 images of the first 100 classes for training and 5,924 images of the other 100 classes for testing without bounding box information.

- CARS196 [16]: CARS196 contains 16,185 images of cars in 196 classes. We use 8,054 images of the first 98 classes for training and 8,131 images of the other 98 classes for testing without bounding box information.

- Standford Online Products (SOP) [21]: SOP contains 120,053 images of products in 22,634 classes. We use 59,551 images of the 11,318 classes for training and 60,502 images of the other 11,316 classes for testing.

### C.2. Implementation

We implement all models using the PyTorch framework [23], and experiments are performed on Nvidia V100 GPUs. For the *conventional evaluation*, we follow the widely used training and testing protocol as [21, 25, 11, 31]. For the *Metric Learning Reality Check (MLRC) evaluation*, we follow the training and evaluation procedure defined in [20].

#### C.2.1 Conventional Evaluation

Input images are augmented by random cropping and horizontal flipping in the training phase, whereas they are center-cropped in the test phase. The size of the cropped images is $224 \times 224$. For the backbone network, the Inception network with batch normalization (BN-Inception) [9] pre-trained with ImageNet [1] is used. We use a global average pooling followed by a fully connected layer for dimensionality reduction and set the dimension of the embedding feature to 512. We freeze batch normalization for CUB200 and CARS196 and keep batch normalization training for SOP by following [24, 31, 11]. The batch size is set to 128 for every experiment. Optimization is performed using Adam optimizer [14] with a learning rate of $10^{-4}$ for CUB200 and CARS196, and $10^{-3}$ for SOP. The learning rate is decayed by a factor of 0.1 at the 50th epoch

for CARS196, and the 20th epoch for CUB200 and SOP. For MemVir, we use warm-up epoch $U_e = 50$ for CARS196 and SOP, and $U_e = 20$ for CUB200 without learning rate decay. With the same hyper-parameters of the baselines, we tune hyper-parameters $N$ and $M$ for MemVir via hyper-parameter search as described in A.

**Proxy-anchor:** For more details of Proxy-anchor loss in terms of implementation, we have found that proxy-anchor loss has been implemented with additional tricks, which is also mentioned in [36]. The additional tricks are as follows: 1) an AdamW optimizer [17] instead of Adam optimizer [14], 2) a parameter warm-up strategy for better optimization stability, 3) instead of an average pooling, a combination of an average and a max pooling following the backbone network. For a fair comparison, we discard those tricks and follow the conventional metric learning protocol in every experiment. Exceptionally, we use the parameter warm-up with one epoch for SOP dataset because the training with Proxy-anchor loss fails without the parameter warm-up strategy.

**Multi-Similarity loss:** In Multi-Similarity (MS) loss [31], we have found that the best scores reported in the paper are conducted with either too small or large batch size, such as a batch size of 80 for CUB200 and batch size of 1000 for SOP. For a fair comparison, we conduct experiments of MS loss with the conventional batch size of 128., including the number of instances of 4. Note that we use the number of instances of 4 instead of 5 from the paper because 128 is not divisible by 5.

### C.2.2 MLRC Evaluation

Each image is resized to make its shorter side to be the length of 256, then augmented by random cropping to have a size between 40 and 256, and by aspect ratio between 3/4 and 4/3 in the training phase. The resulting image is then resized to $227 \times 227$ and flipped horizontally with a 50% probability. In the test phase, each image is resized to 256 and center-cropped to 227. We use BN-Inception for the backbone network with an output embedding size of 128. Optimization is performed using RMSprop optimizer with a learning rate of $10^{-6}$ and a batch size of 32. To find the best hyper-parameters for loss functions, we run 50 experiments of hyper-parameter search with 4-fold cross-validation of each experiment. With the best hyper-parameters found, we conduct 10 training runs and report the average and confidence intervals to be less subject to random seed noise. We report both separated (128-dim) and concatenated (512-dim) performance, where the 512-dim embedding is concatenated and $l_2$-normalized of 128-dim embedding of the 4 models.

### C.3. fANOVA

We apply the fANOVA [8] analysis framework to estimate the impact of each hyper-parameter on the performance of MemVir in Section 4.4 and D.5. fANOVA predicts the marginal performance using a predictive model (random forest), which is a function of the model's hyper-parameters. Then, it determines the extent to which each hyper-parameter or pair-wise interaction contributes to the model performance. In the experiments of MemVir, we conduct 5 training runs for each pair of $(N, M)$, and each pair is created by the combination of range $N$ and range $M$. For CUB200 and CAS196, range $N$ is 5 to 50 in 5 intervals including 1, and range $M$ is 0 to 100 in 10 intervals. For SOP, we reduce the range $N$ to be 1 to 7 because of memory limitation of one gpu. These experimental results are used to train random forests for fANOVA analysis.

## D. Extended Experiments

### D.1. Analysis of Memory and Computational Cost

In this section, we analyze memory and computational cost of MemVir with the same experimental setting described in Section C.2.1. MemVir requires $O(BNM(D + C))$ for the memory queues, $O(BCN^2)$ for the similarity matrix, and $O(BCN^2)$ for the computational complexity during the training phase, where $B$, $C$, and $D$ are batch size, number of classes, and feature dimension, respectively. In the inference phase, MemVir requires no additional memory or computational cost. As shown in Figure B, the memory usage of MemVir with 128 batch size increases as $N$ and $M$ are increased. Compared to the Norm-softmax model, MemVir(1,100) and MemVir(45,10) improve $+1.3\%$ and $+3.5\%$ performances with additional $52MB$ and $704MB$ GPU memory, respectively. MemVir(50,100), which increases the number of classes and embeddings by 50 times, only requires additional 2.9GB GPU memory and shows better memory efficiency than the baseline with 256 batch size, which requires 6.8GB more GPU memory than MemVir(50,100) with 128 batch size. Even though the memory usage of MemVir would be larger for datasets with a large number of classes, it can be controlled by placing a reduced number of class weights in $\mathbb{W}$, which are corresponding classes of current step embeddings. In terms of performance, applying MemVir is more effective than increasing the batch size, which is empirically shown in Section 4.3.

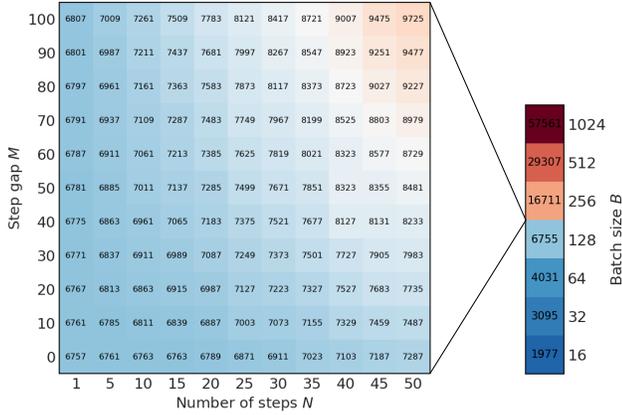

(a) *MemVir(N,M)* + Norm-softmax with 128 batch size  (b) Norm-softmax

Figure B. Memory usage (MB) of MemVir($N$,$M$) + Norm-softmax with 128 batch size and Norm-softmax with different batch size on CARS196 dataset.

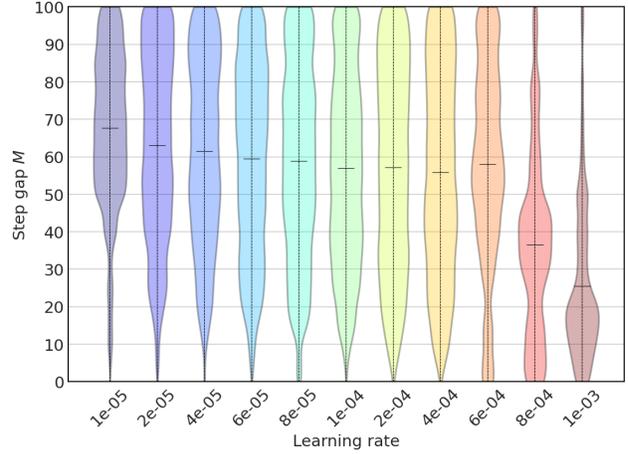

Figure C. Distributions of relative performances for each learning rate by step gap $M$. For each learning rate, Recall@1 performances of MemVir(1,$M$) on CARS196 dataset are normalized across step gap $M$.

## D.2. Impact of Learning Rate

Typically, memory-based methods [13, 33] get influenced by learning rate because it affects the difference between training steps. As discussed in Section 3.2, MemVir can control the difference between training steps with margin parameter $M$. To see the impact of learning rate, we train MemVir(1,M) with a range $M$ from 0 to 100 in 5 intervals and normalize Recall@1 performances across step gap $M$ for each learning rate. We plot the distributions of relative performances by step gap $M$ for each learning rate. As described in Figure C, the larger learning rate requires the smaller step gap $M$ for better performance. This is because a sufficient difference between the training steps can be achieved by a large gap $M$ for a small learning rate and a small gap $M$ for a large learning rate. Thus, the different learning rates can be controlled by the step gap $M$ for MemVir.

## D.3. Impact of Warm-up

As discussed in Section 3.3.1, the warm-up period of MemVir enables the model to avoid training with distractive virtual classes from the initial step. To see the impact of the warm-up period, we conduct experiments by differentiating the warm-up epoch with MemVir(1,100) and MemVir(50,10). As shown in Figure D, MemVir(1,100) shows the lowest performance when $U_e = 0$ and stable performance for $U_e > 0$. For MemVir(50,10), the lowest performance is also shown when $U_e = 0$ and the performance increases until $U_e = 60$, then decreases. It is noteworthy that the lowest performance of MemVir with $U_e = 0$ still shows higher performance than that of the baseline. The results show that proper steps of warm-up help increase the capability of MemVir, and this pattern stands out more to MemVir with longer scheduling of virtual classes addition.

## D.4. Robustness to Input Deformation

We now evaluate the quality of representations learned with MemVir with respect to generalization to input deformations. We train models with Norm-softmax loss and MemVir + Norm-softmax loss on CARS196 dataset but test them on the novel (not seen during training) input deformations of the test set. For the input deformation, we use the *imgaug* [10] python library and the details of deformations are as follows.

- *Cutout*: Each image is randomly filled with two gray pixels that are 20% of the image size.

- *Dropout*: $p$% of pixels are dropped from each image, where $p$ is randomly sampled from a range $0\% \leq p \leq 20\%$.

- *Zoom-in* and *zoom-out*: Each image is transformed by zoom-in and zoom-out with scale of 50% and 150%, respectively.

- *Rotation* and *shearing*: Each image is transformed by rotation and shearing with a randomly sampled degree between -30° and 30°.

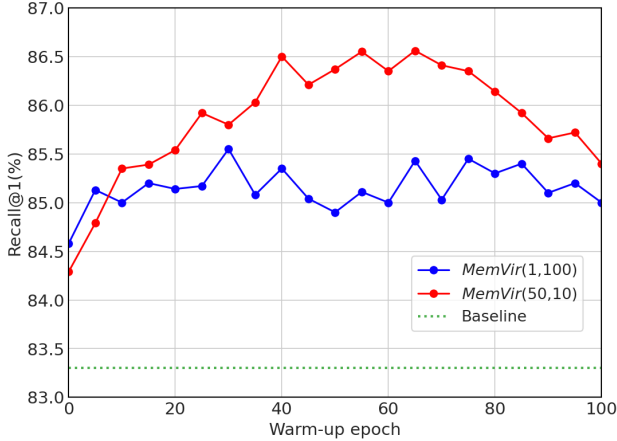

Figure D. Impact of warm-up epoch $U_e$ of MemVir + Norm-softmax on CARS196 dataset. We report the performance of MemVir with different warm-up epoch. Note that the performance of baseline Norm-softmax is unrelated to the warm-up epoch.

| Deformation | Norm-softmax | *MemVir* |
|---|---|---|
| W/o deformation | 83.3 | 86.6 (+3.3) |
| Cutout | 75.3 | 79.2 (+3.9) |
| Dropout | 59.7 | 67.7 (+8.0) |
| Zoom in | 64.3 | 68.1 (+3.9) |
| Zoom out | 78.3 | 81.7 (+3.4) |
| Rotation | 70.8 | 73.4 (+2.6) |
| Shearing | 70.3 | 73.2 (+3.0) |
| Gaussian noise | 65.1 | 71.2 (+6.1) |
| Gaussian blur | 74.4 | 78.2 (+3.8) |

Table B. Recall@1(%) performance of input deformations with CARS196 trained models. We compare Norm-softmax with MemVir(50,10) + Norm-softmax.

| (N,M) | (0,0)-baseline | (1,100) | (5,100) | (10,100) | (20,100) |
|---|---|---|---|---|---|
| Only-weights | 83.6 | 84.4 | **84.5** | 84.3 | 83.8 |
| MemVir | | 84.9 | 85.1 | 85.5 | **85.8** |

Table C. Recall@1 (%) of only-weights and MemVir on CARS196.

- *Gaussian noise*: Gaussian noise is applied to each image, where the noise is sampled per pixel from a normal distribution $N(0, s)$ and $s$ is sampled between 0 and $0.2 \times 255$.

- *Gaussian blur*: Gaussian kernel with a sigma of 3.0 is applied to each image.

As shown in Table B, performances of Norm-softmax are degraded significantly when applied input deformations. On the other hand, MemVir exhibits relatively smaller performance degradation compared to that of the Norm-softmax, and shows better robustness to all input deformations, particularly for dropout and gaussian noise. This demonstrates that MemVir allows the model to obtain a more generalized embedding space.

### D.5. Impact of Hyper-parameters

In addition to the hyper-parameter analysis of CARS196 in Section 4.4, we include extra analyses on CUB200 and SOP. As shown in Figure E, the performance on CUB200 increases until the margin $M = 5$, and then decreases. For the margin, the performance improves as the margin $M$ increases. In the case of SOP, the performance increases as both $N$ and $M$ increase while there is a slight degradation of performance around $M = 10$. As mentioned in Section 4.4, the performance pattern and importance of each hyper-parameter differs for each dataset because of different data characteristics. However, we can observe that, for both CUB200 and SOP, the best performance is achieved when $N$ and $M$ are larger than 1 and 0, respectively.

### D.6. Impact of Embeddings and Class Weights in Virtual Class

To investigate the quantitative impact of embeddings and class weights in virtual classes, we conduct an experiment by using only class weights in virtual classes. Note that using only embeddings in virtual classes is not possible because every embedding requires corresponding class weights in loss computation. Table C shows that using only class weights increases performance than the baseline, but using both embeddings and class weights (MemVir) outperforms it. It suggests the necessity of embeddings to form proper virtual classes.

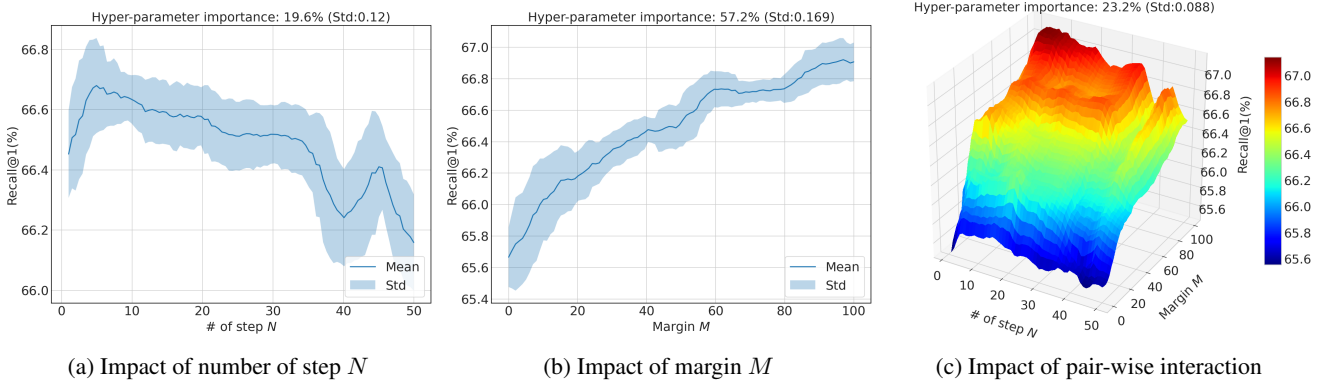

Figure E. Impact of hyper-parameters on CUB200 with fANOVA analysis framework.

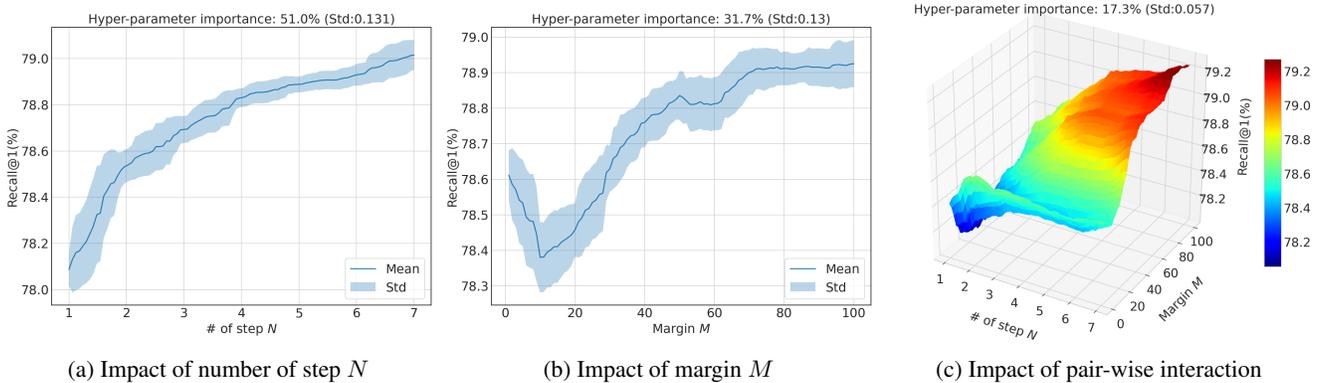

Figure F. Impact of hyper-parameters on SOP with fANOVA analysis framework.

### D.7. Comparison with Related Methods

We provide an extended comparison with related methods from image recognition tasks following Section 4.5. As shown in Table Da, we set two different experimental setups to compare the methods in various experimental settings. Setup 1 is following the experimental settings from BroadFace [13] and CurricularFace [7], and setup 2 is from the conventional DML settings described in Section C.2.1 with ResNet50 [6] backbone. As XBM [33] shares the memory-based idea with BroadFace, we conduct experiments of XBM in ArcFace loss to compare with BroadFace. As shown in Table D, Virtual softmax degrades the performance for both setups 1 and 2, whereas MemVir improves the performance for both setups. In Virtual softmax, the single virtual class weight is created by $W_{virt} = \frac{\|W_{y_i}\|X_i}{\|X_i\|}$. We observe that the logit with virtual class $W_{virt}^T X_i$ is much larger than the positive logit $W_{y_i}^T X_i$ due to the same direction of vectors $W_{virt}$ and $X_i$, and it distracts the model from stable training. Combining XBM with ArcFace shows a little performance improvement for both setups, but we observe performance degradation when the memory size is large, as reported in BroadFace. To resolve this problem, BroadFace presents a compensation technique and gradient control. For the details, the compensation technique compensates memorized embeddings by considering class weights updating, while the gradient control computes the loss function into two ways: (1) loss from a mini-batch is for updating the backbone network. (2) loss from the mini-batch and past embeddings is for updating the class weights. BroadFace shows a higher performance boost with the SGD optimizer in setup 1 than the Adam optimizer in setup 2. We observe that BroadFace is sensitive to optimizer type, which could be because of the specifically designed gradient control. On the other hand, MemVir achieves larger performance gains for both setups without any modification of loss functions. CurricularFace shows competitive performance in both setups with an embedded curriculum learning process. When MemVir is applied, the performance of CurricularFace could further be improved by exploiting augmented information from virtual classes.

|  | Backbone | Dimension | Batch size | Optimizer | Initial LR |
| --- | --- | --- | --- | --- | --- |
| Setup 1 | ResNet50 | 512 | 512 | SGD | 0.005 |
| Setup 2 | ResNet50 | 512 | 128 | Adam | 0.0001 |

(a) Two different experimental setups.

| | CARS196 | | | | SOP | | | |
| --- | --- | --- | --- | --- | --- | --- | --- | --- |
| Method | R@1 | R@2 | R@4 | R@8 | R@1 | R@10 | R@100 | R@1000 |
| Softmax | 78.3 | 86.4 | 91.9 | **95.7** | 76.6 | 89.4 | 95.8 | **98.8** |
| Virtual Softmax | 75.1 | 84.1 | 90.1 | 94.0 | 74.5 | 87.9 | 94.8 | 98.3 |
| *MemVir* + Softmax | **79.2** | **87.0** | **92.1** | 95.7 | **78.9** | **90.6** | **96.2** | 98.8 |
| ArcFace | 78.8 | 86.4 | 91.7 | 95.4 | 76.9 | 89.1 | 95.0 | 98.2 |
| XBM + ArcFace | 78.9 | 86.2 | 91.9 | 95.5 | 78.1 | 89.7 | 95.8 | 98.2 |
| BroadFace + ArcFace | 79.5 | 87.3 | 92.0 | 95.5 | 80.2 | 91.0 | 95.9 | 98.4 |
| *MemVir* + ArcFace | **80.7** | **88.1** | **92.7** | **95.8** | **80.8** | **91.3** | **96.5** | **98.9** |
| CurricularFace | 79.9 | 87.3 | 92.0 | 95.6 | 79.8 | 90.7 | 95.6 | 98.2 |
| *MemVir* + CurricularFace | **81.0** | **87.9** | **92.9** | **95.8** | **81.3** | **91.7** | **96.5** | **98.8** |

(b) Performance (%) comparison in experimental setup 1.

| | CARS196 | | | | SOP | | | |
| --- | --- | --- | --- | --- | --- | --- | --- | --- |
| Method | R@1 | R@2 | R@4 | R@8 | R@1 | R@10 | R@100 | R@1000 |
| Softmax | 82.4 | 88.8 | 92.1 | 95.2 | 78.1 | 89.1 | 95.2 | 97.9 |
| Virtual Softmax | 77.6 | 85.3 | 90.7 | 94.3 | 77.3 | 87.8 | 94.1 | 97.2 |
| *MemVir* + Softmax | **86.8** | **92.3** | **95.5** | **97.6** | **78.9** | **90.6** | **96.2** | **98.8** |
| ArcFace | 83.2 | 89.5 | 93.7 | 96.4 | 78.6 | 90.3 | 95.8 | 98.5 |
| XBM + ArcFace | 83.9 | 90.3 | 94.2 | 96.5 | 78.8 | 90.1 | 96.0 | 98.3 |
| BroadFace + ArcFace | 83.9 | 90.5 | 94.2 | **96.9** | 79.1 | 90.7 | **96.2** | **98.8** |
| *MemVir* + ArcFace | **85.4** | **90.9** | **94.7** | **96.9** | **79.5** | **90.9** | **96.2** | **98.8** |
| CurricularFace | 83.5 | 90.1 | 94.2 | 96.5 | 78.5 | 90.1 | 95.6 | 98.4 |
| *MemVir* + CurricularFace | **85.4** | **91.0** | **94.5** | **96.9** | **79.3** | **90.5** | **95.8** | **98.5** |

(c) Performance (%) comparison in experimental setup 2.

Table D. Performance (%) comparison with related methods in two different experimental setups.

## D.8. Visualization of Embedding Space

For further understanding, we include extended t-SNE [18] visualization of embedding space, followed by Figure 6 in Section 4.2. As illustrated in Figure Ga, Gb, and Gc, embedding features from the seen classes are getting clustered by training process, and further training may cause a model overly fitted to the seen classes. To alleviate this strong focus on seen classes, MemVir begins to add virtual classes slowly and increases learning difficulty gradually, as described in Figure Gd, Ge, Gf, Gg and Gh. Actual and virtual classes, which are originated from the same class label, tend to be located close to each other. Further training allows the model to learn how to discriminate additional classes effectively, as illustrated in Figure Gi, Gj, Gk, and Gl. Finally, we obtain a model with sufficient discriminative power over all actual and virtual classes at the 200th epoch.

## D.9. Comparison with State-of-the-art

This section includes extended comparison with state-of-the-art methods for both *conventional evaluation* and *MLRC evaluation* in addition to Section 4.6. As shown in Table E, G, and I of conventional evaluation, we report additional Recall@k performance and comparison with different types of DML methods. For MLRC evaluation, we report both separated (128-dim) and concatenated (512-dim) performance as presented in Table F, H, and J. In conventional evaluation, MemVir shows a significant performance boost in all Recall@k for every dataset and loss function. Compared with different types of DML methods, including ensemble, sample generation, memory-based, pair-based, proxy-based, and softmax variants,

◊ : Class weight,  Embedding color (step): • (current) → • (-1(M+1)) → • (-2(M+1)) → • (-3(M+1)) → • (-4(M+1)) → • (-5(M+1))

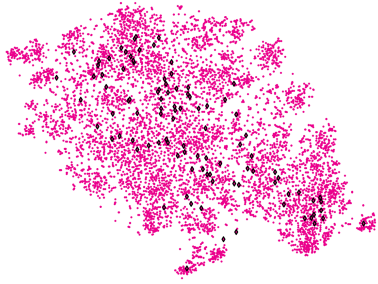
(a) 1st epoch, # of classes = C

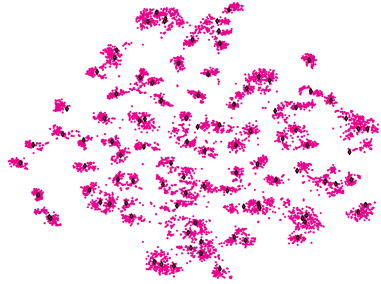
(b) 25th epoch, # of classes = C

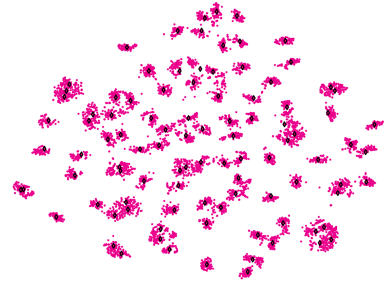
(c) 50th epoch, # of classes = C

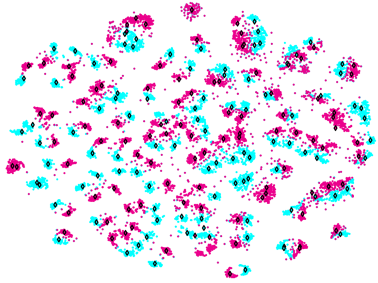
(d) 52nd epoch, # of classes = 2C

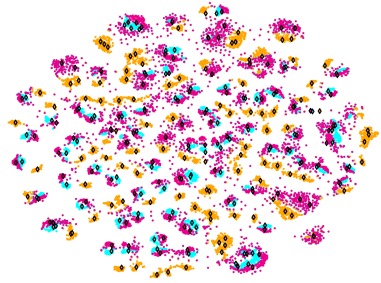
(e) 54th epoch, # of classes = 3C

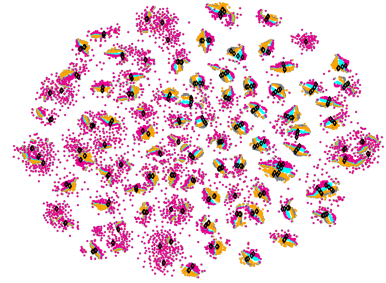
(f) 56th epoch, # of classes = 4C

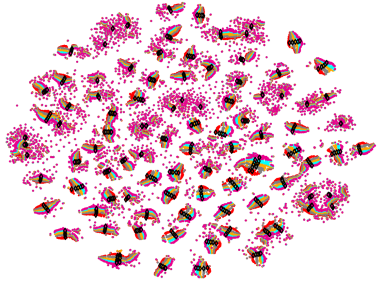
(g) 58th epoch, # of classes = 5C

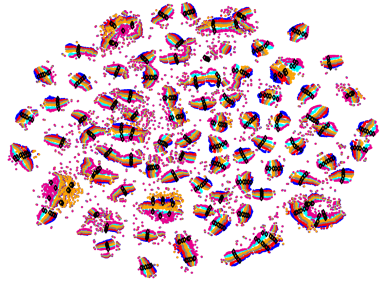
(h) 60th epoch, # of classes = 6C

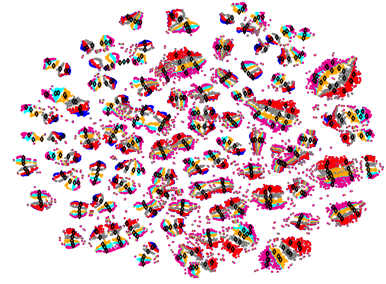
(i) 90th epoch, # of classes = 6C

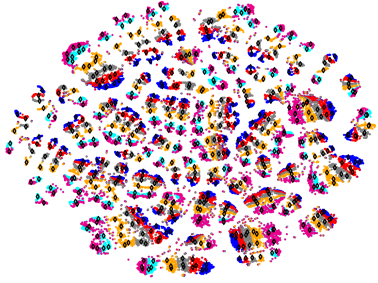
(j) 120th epoch, # of classes = 6C

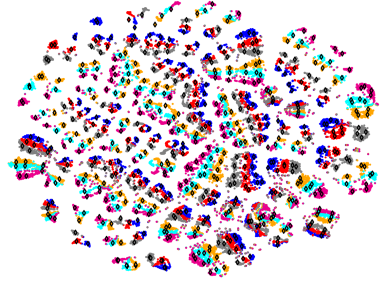
(k) 150th epoch, # of classes = 6C

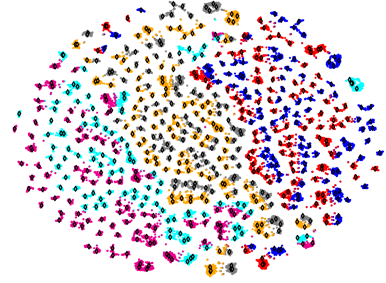
(l) 200th epoch, # of classes = 6C

Figure G. t-SNE visualization of 512-dimensional embedding space. Embedding features are extracted by a model trained with MemVir(5,100) on CARS196 training data. Each color indicates a step for embedding features.

MemVir shows competitive performance for all datasets. In MLRC evaluation, MemVir enjoys a high-performance gain over every dataset and loss function in both separated (128-dim) and concatenated (512-dim) experiments. Considering MLRC evaluation is designed for fair evaluation, the results demonstrate that MemVir is a flexible and powerful training strategy for many existing softmax variants and proxy-based losses.

| T | Method | Net | Dim | R@1 | | R@2 | | R@4 | | R@8 | |
|---|---|---|---|---|---|---|---|---|---|---|---|
| | | | | CUB-200-2011 [28] | | | | | | | |
| Ens | HDC [34] | G | 384 | 53.6 | - | 65.7 | - | 77.0 | - | 85.6 | - |
| Ens | A-BIER [22] | G | 512 | 57.5 | - | 68.7 | - | 78.3 | - | 86.2 | - |
| Ens | ABE [12] | G | 512 | 60.6 | - | 71.5 | - | 79.8 | - | 87.4 | - |
| Gen | DAML [3] + N-pair | G | 512 | 52.7 | - | 65.4 | - | 75.5 | - | 84.3 | - |
| Gen | HDML [35] + N-pair | G | 512 | 53.7 | - | 65.7 | - | 76.7 | - | 85.7 | - |
| Gen | Symm [5] + N-pair | G | 512 | 55.9 | - | 67.6 | - | 78.3 | - | 86.2 | - |
| Gen | EE [15] + MS | G | 512 | 57.4 | - | 68.7 | - | 79.5 | - | 86.9 | - |
| Gen | Symm [5] + MS | BN | 512 | 64.9 | - | 76.4 | - | 84.6 | - | 90.5 | - |
| Gen | EE [15] + MS | BN | 512 | 65.1 | - | 76.8 | - | 86.1 | - | 91.0 | - |
| M | XBM [33] + Contrastive | BN | 512 | 65.8 | - | 75.9 | - | 84.0 | - | 89.9 | - |
| Pair | HTL [4] | BN | 512 | 57.1 | - | 68.8 | - | 78.7 | - | 86.5 | - |
| Pair | RLL-H [32] | BN | 512 | 57.4 | - | 69.7 | - | 79.2 | - | 86.9 | - |
| Pair | Multi-Similarity (MS)[†] [31] | BN | 512 | 64.5 | - | 76.2 | - | 84.6 | - | 90.5 | - |
| Softmax variant / Proxy | SoftTriple [24] | BN | 512 | 65.4 | - | 76.4 | - | 84.5 | - | 90.4 | - |
| Softmax variant / Proxy | ProxyGML [36] | BN | 512 | 66.6 | - | 77.6 | - | 86.4 | - | - | - |
| Softmax variant / Proxy | Circle [26] | BN | 512 | 66.7 | - | 77.4 | - | 86.2 | - | 91.2 | - |
| Softmax variant / Proxy | Softmax | BN | 512 | 64.2 | - | 75.7 | - | 84.1 | - | 89.9 | - |
| Softmax variant / Proxy | *MemVir* + Softmax | BN | 512 | 66.8 | (+2.6) | 76.9 | (+1.2) | 85.4 | (+1.3) | 91.2 | (+1.3) |
| Softmax variant / Proxy | Norm-softmax [29] | BN | 512 | 64.9 | - | 75.7 | - | 84.3 | - | 90.5 | - |
| Softmax variant / Proxy | *MemVir* + Norm-softmax | BN | 512 | 67.3 | (+2.4) | 77.2 | (+1.5) | 85.3 | (+1.0) | 90.8 | (+0.3) |
| Softmax variant / Proxy | Cosface [30] | BN | 512 | 65.7 | - | 76.2 | - | 84.7 | - | 90.6 | - |
| Softmax variant / Proxy | *MemVir* + Cosface | BN | 512 | 67.7 | (+2.0) | 77.8 | (+1.6) | 85.7 | (+1.0) | 91.1 | (+0.5) |
| Softmax variant / Proxy | Arcface [2] | BN | 512 | 66.1 | - | 76.6 | - | 84.8 | - | 90.7 | - |
| Softmax variant / Proxy | *MemVir* + Arcface | BN | 512 | 67.4 | (+1.3) | 77.7 | (+1.1) | 85.5 | (+0.7) | 91.2 | (+0.5) |
| Softmax variant / Proxy | Proxy-NCA [19] | BN | 512 | 64.3 | - | 75.3 | - | 83.6 | - | 89.6 | - |
| Softmax variant / Proxy | *MemVir* + Proxy-NCA | BN | 512 | 68.3 | (+4.0) | 78.9 | (+3.6) | 85.7 | (+2.1) | 90.9 | (+1.3) |
| Softmax variant / Proxy | Proxy-anchor[†] [11] | BN | 512 | 67.7 | - | 78.5 | - | 85.7 | - | 90.9 | - |
| Softmax variant / Proxy | *MemVir* + Proxy-anchor | BN | 512 | **69.0** | (+1.3) | **79.2** | (+0.7) | **86.8** | (+1.1) | **91.6** | (+0.7) |
| - | Average boost | - | - | - | (+2.3) | - | (+1.6) | - | (+1.2) | - | (+0.8) |
| - | Minimum boost | - | - | - | (+1.3) | - | (+0.7) | - | (+0.7) | - | (+0.3) |
| - | Maximum boost | - | - | - | (+4.0) | - | (+3.6) | - | (+2.1) | - | (+1.3) |

Table E. **[Conventional evaluation]** Recall@k (%) on CUB-200-2011 dataset in image retrieval task. Method type (T) is denoted by abbreviations (*Ens*: ensemble, *Gen*: sample generation, *M*: memory-based, *Pair*: pair-based losses, *Softmax variant / Proxy*: softmax variants and proxy-based losses). Backbone network (Net) also is denoted by abbreviations (*G*: GoogleNet [27], *BN*: BN-Inception [9]). [†] denotes evaluation in a fair setting described in Section C.2.1.

| CUB-200-2011 [28] | Concatenated (512-dim) | | | Separated (128-dim) | | |
|---|---|---|---|---|---|---|
| Loss | P@1 | RP | MAP@R | P@1 | RP | MAP@R |
| Norm-softmax [29] | 65.65 ± 0.30 | 35.99 ± 0.15 | 25.25 ± 0.13 | 58.75 ± 0.19 | **31.75 ± 0.12** | **20.96 ± 0.11** |
| *MemVir* + Norm-softmax | **69.22 ± 0.15** | **37.92 ± 0.16** | **27.10 ± 0.13** | **59.83 ± 0.23** | 31.46 ± 0.16 | 20.55 ± 0.14 |
| CosFace [30] | 67.32 ± 0.32 | 37.49 ± 0.21 | 26.70 ± 0.23 | 59.63 ± 0.36 | **31.99 ± 0.22** | 21.21 ± 0.22 |
| *MemVir* + CosFace | **69.79 ± 0.26** | **37.85 ± 0.23** | **27.08 ± 0.28** | **61.33 ± 0.30** | **31.99 ± 0.18** | **21.30 ± 0.17** |
| ArcFace [2] | 67.50 ± 0.25 | 37.31 ± 0.21 | 26.45 ± 0.20 | 60.17 ± 0.32 | 32.37 ± 0.17 | 21.49 ± 0.16 |
| *MemVir* + ArcFace | **69.33 ± 0.41** | **37.82 ± 0.28** | **26.96 ± 0.25** | **61.38 ± 0.23** | **32.53 ± 0.13** | **21.58 ± 0.12** |
| Proxy-NCA [19] | 65.69 ± 0.43 | 35.14 ± 0.26 | 24.21 ± 0.27 | 57.88 ± 0.30 | 30.16 ± 0.22 | 19.32 ± 0.21 |
| *MemVir* + Proxy-NCA | **69.25 ± 0.32** | **37.31 ± 0.12** | **26.43 ± 0.17** | **60.08 ± 0.25** | **31.26 ± 0.15** | **20.30 ± 0.14** |
| Proxy-anchor [11] | 69.73 ± 0.31 | 38.23 ± 0.37 | 27.44 ± 0.35 | 61.50 ± 0.34 | 32.94 ± 0.25 | 22.19 ± 0.25 |
| *MemVir* + Proxy-anchor | **69.81 ± 0.28** | **38.57 ± 0.14** | **27.83 ± 0.16** | **62.58 ± 0.28** | **33.69 ± 0.18** | **22.75 ± 0.16** |

Table F. **[MLRC evaluation]** Performance (%) on CUB-200-2011 dataset in image retrieval task. We report the performance of concatenated 512-dim and separated 128-dim. Bold numbers indicate the best score within the same loss.

| T | Method | Net | Dim | R@1 | | R@2 | | R@4 | | R@8 | |
|---|---|---|---|---|---|---|---|---|---|---|---|
| | | | | CARS196 [16] | | | | | | | |
| Ens | HDC [34] | G | 384 | 73.7 | - | 83.2 | - | 89.5 | - | 93.8 | - |
| | A-BIER [22] | G | 512 | 82.0 | - | 89.0 | - | 93.2 | - | 96.1 | - |
| | ABE [12] | G | 512 | 85.2 | - | 90.5 | - | 94.0 | - | 96.1 | - |
| Gen | DAML [3] + N-pair | G | 512 | 75.1 | - | 83.8 | - | 89.7 | - | 93.5 | - |
| | HDML [35] + N-pair | G | 512 | 79.1 | - | 87.1 | - | 92.1 | - | 95.5 | - |
| | Symm [5] + N-pair | G | 512 | 76.5 | - | 84.3 | - | 90.4 | - | 94.1 | - |
| | EE [15] + MS | G | 512 | 76.1 | - | 84.2 | - | 89.8 | - | 93.8 | - |
| | Symm [5] + MS | BN | 512 | 82.4 | - | 89.2 | - | 93.3 | - | 96.1 | - |
| | EE [15] + MS | BN | 512 | 82.7 | - | 89.2 | - | 93.8 | - | 96.4 | - |
| M | XBM [33] + Contrastive | BN | 512 | 82.0 | - | 88.7 | - | 93.1 | - | 96.1 | - |
| Pair | HTL [4] | BN | 512 | 81.4 | - | 88.0 | - | 92.7 | - | 95.7 | - |
| | RLL-H [32] | BN | 512 | 74.0 | - | 83.6 | - | 90.1 | - | 94.1 | - |
| | Multi-Similarity (MS)[†] [31] | BN | 512 | 82.1 | - | 88.8 | - | 93.2 | - | 96.1 | - |
| Softmax variant / Proxy | SoftTriple [24] | BN | 512 | 84.5 | - | 90.7 | - | 94.5 | - | 96.9 | - |
| | ProxyGML [36] | BN | 512 | 85.5 | - | 91.8 | - | 95.3 | - | - | - |
| | Circle [26] | BN | 512 | 83.4 | - | 89.8 | - | 94.1 | - | 96.5 | - |
| | Softmax | BN | 512 | 81.5 | - | 89.0 | - | 93.6 | - | 96.8 | - |
| | *MemVir* + Softmax | BN | 512 | 86.5 | (+5.0) | 92.4 | (+3.4) | 95.6 | (+2.0) | 97.4 | (+0.6) |
| | Norm-softmax [29] | BN | 512 | 83.3 | - | 89.7 | - | 94.1 | - | 96.7 | - |
| | *MemVir* + Norm-softmax | BN | 512 | **86.8** | (+3.5) | 92.3 | (+2.6) | 95.4 | (+1.3) | 97.4 | (+0.7) |
| | Cosface [30] | BN | 512 | 83.6 | - | 89.9 | - | 94.2 | - | 96.6 | - |
| | *MemVir* + Cosface | BN | 512 | 86.6 | (+3.0) | 91.8 | (+1.9) | 95.1 | (+0.9) | 97.3 | (+0.7) |
| | Arcface [2] | BN | 512 | 83.7 | - | 90.0 | - | 94.3 | - | 96.8 | - |
| | *MemVir* + Arcface | BN | 512 | 86.5 | (+2.8) | 91.9 | (+1.9) | 95.1 | (+0.8) | 97.1 | (+0.3) |
| | Proxy-NCA [19] | BN | 512 | 82.0 | - | 89.2 | - | 93.8 | - | 96.4 | - |
| | *MemVir* + Proxy-NCA | BN | 512 | 86.5 | (+4.5) | 91.8 | (+2.6) | 95.5 | (+1.7) | 97.4 | (+1.0) |
| | Proxy-anchor[†] [11] | BN | 512 | 84.9 | - | 91.1 | - | 94.6 | - | 96.9 | - |
| | *MemVir* + Proxy-anchor | BN | 512 | 86.7 | (+1.8) | 92.0 | (+0.9) | 95.2 | (+0.6) | 97.4 | (+0.5) |
| - | Average boost | - | - | - | (+3.4) | - | (+2.2) | - | (+1.2) | - | (+0.6) |
| | Minimum boost | - | - | - | (+1.8) | - | (+0.9) | - | (+0.6) | - | (+0.3) |
| | Maximum boost | - | - | - | (+5.0) | - | (+3.4) | - | (+2.0) | - | (+1.0) |

Table G. [**Conventional evaluation**] Recall@k (%) on CARS196 dataset in image retrieval task. Method type (T) is denoted by abbreviations (*Ens*: ensemble, *Gen*: sample generation, *M*: memory-based, *Pair*: pair-based losses, *Softmax variant / Proxy*: softmax variants and proxy-based losses). Backbone network (Net) also is denoted by abbreviations (*G*: GoogleNet [27], *BN*: BN-Inception [9]). [†] denotes evaluation in a fair setting described in Section C.2.1.

| CARS196 [16] | Concatenated (512-dim) | | | Separated (128-dim) | | |
|---|---|---|---|---|---|---|
| Loss | P@1 | RP | MAP@R | P@1 | RP | MAP@R |
| Norm-softmax [29] | 83.16 ± 0.25 | 36.20 ± 0.26 | 26.00 ± 0.30 | 72.55 ± 0.18 | 29.35 ± 0.20 | 18.73 ± 0.20 |
| *MemVir* + Norm-softmax | **85.81 ± 0.18** | **38.78 ± 0.19** | **28.92 ± 0.17** | **76.01 ± 0.23** | **30.86 ± 0.16** | **20.36 ± 0.16** |
| CosFace [30] | 85.52 ± 0.24 | 37.32 ± 0.28 | 27.57 ± 0.30 | 74.67 ± 0.20 | 29.01 ± 0.11 | 18.80 ± 0.12 |
| *MemVir* + CosFace | **87.57 ± 0.13** | **39.10 ± 0.21** | **29.56 ± 0.26** | **76.86 ± 0.28** | **30.59 ± 0.22** | **20.23 ± 0.24** |
| ArcFace [2] | 85.44 ± 0.28 | 37.02 ± 0.29 | 27.22 ± 0.30 | 72.10 ± 0.37 | 27.29 ± 0.17 | 17.11 ± 0.18 |
| *MemVir* + ArcFace | **88.02 ± 0.18** | **39.12 ± 0.15** | **29.63 ± 0.15** | **78.58 ± 0.26** | **31.03 ± 0.25** | **20.89 ± 0.26** |
| Proxy-NCA [19] | 83.56 ± 0.27 | 35.62 ± 0.28 | 25.38 ± 0.31 | 73.46 ± 0.23 | 28.90 ± 0.22 | 18.29 ± 0.22 |
| *MemVir* + Proxy-NCA | **87.02 ± 0.15** | **38.51 ± 0.15** | **28.76 ± 0.16** | **76.35 ± 0.28** | **30.29 ± 0.23** | **19.81 ± 0.25** |
| Proxy-anchor [11] | 86.20 ± 0.21 | 39.08 ± 0.31 | 29.37 ± 0.29 | 76.97 ± 0.40 | 31.71 ± 0.53 | 21.29 ± 0.56 |
| *MemVir* + Proxy-anchor | **86.40 ± 0.18** | **40.27 ± 0.20** | **30.58 ± 0.20** | **76.97 ± 0.26** | **32.87 ± 0.21** | **22.31 ± 0.22** |

Table H. [**MLRC evaluation**] Performance (%) on CARS196 dataset in image retrieval task. We report the performance of concatenated 512-dim and separated 128-dim. Bold numbers indicate the best score within the same loss.

| T | Method | Net | Dim | R@1 | | R@10 | | R@100 | | R@1000 | |
|---|---|---|---|---|---|---|---|---|---|---|---|
| | | | | Stanford Online Products [21] | | | | | | | |
| Ens | HDC [34] | G | 384 | 69.5 | - | 84.4 | - | 92.8 | - | 97.7 | - |
| | A-BIER [22] | G | 512 | 74.2 | - | 86.9 | - | 94.0 | - | 97.8 | - |
| | ABE [12] | G | 512 | 76.3 | - | 88.4 | - | 94.8 | - | 98.2 | - |
| Gen | DAML [3] + N-pair | G | 512 | 68.4 | - | 83.5 | - | 92.3 | - | - | - |
| | HDM [35] + N-pair | G | 512 | 68.7 | - | 83.2 | - | 92.4 | - | - | - |
| | Symm [5] + N-pair | G | 512 | 73.2 | - | 86.7 | - | 94.8 | - | - | - |
| | EE [15] + MS | G | 512 | 78.1 | - | 90.3 | - | 95.8 | - | - | - |
| | Symm [5] + MS | BN | 512 | 76.9 | - | 89.8 | - | 95.9 | - | 98.8 | - |
| | EE [15] + MS | BN | 512 | 77.0 | - | 89.5 | - | 96.0 | - | 98.8 | - |
| M | XBM + Contrastive [33] | BN | 512 | 79.5 | - | 90.8 | - | 96.1 | - | 98.7 | - |
| Pair | HTL [4] | BN | 512 | 74.8 | - | 88.3 | - | 94.8 | - | 98.4 | - |
| | RLL-H [32] | BN | 512 | 76.1 | - | 89.1 | - | 95.4 | - | - | - |
| | Multi-Similarity (MS)[†] [31] | BN | 512 | 76.3 | - | 89.7 | - | 96.0 | - | 98.8 | - |
| Softmax variant / Proxy | SoftTriple [24] | BN | 512 | 78.3 | - | 90.4 | - | 96.0 | - | 98.3 | - |
| | ProxyGML [36] | BN | 512 | 78.0 | - | 90.6 | - | 96.2 | - | - | - |
| | Circle [26] | BN | 512 | 78.3 | - | 90.5 | - | 96.1 | - | 98.6 | - |
| | Softmax | BN | 512 | 76.3 | - | 88.5 | - | 94.8 | - | 98.1 | - |
| | *MemVir* + Softmax | BN | 512 | 77.8 | (+1.5) | 89.8 | (+1.3) | 95.4 | (+0.6) | 98.4 | (+0.3) |
| | Norm-softmax [29] | BN | 512 | 78.6 | - | 90.5 | - | 96.0 | - | 98.6 | - |
| | *MemVir* + Norm-softmax | BN | 512 | 79.6 | (+1.0) | 90.9 | (+0.4) | 96.1 | (+0.1) | 98.7 | (+0.1) |
| | Cosface [30] | BN | 512 | 78.6 | - | 90.4 | - | 95.8 | - | 98.5 | - |
| | *MemVir* + Cosface | BN | 512 | 79.7 | (+1.1) | 90.5 | (+0.1) | 95.8 | (0.0) | 98.5 | (0.0) |
| | Arcface [2] | BN | 512 | 78.8 | - | 90.5 | - | 95.9 | - | 98.6 | - |
| | *MemVir* + Arcface | BN | 512 | **80.0** | (+1.2) | 90.9 | (+0.4) | 96.1 | (+0.2) | 98.7 | (+0.1) |
| | Proxy-NCA [19] | BN | 512 | 78.1 | - | 90.0 | - | 95.9 | - | 98.7 | - |
| | *MemVir* + Proxy-NCA | BN | 512 | 79.2 | (+1.1) | 90.4 | (+0.4) | 96.0 | (+0.1) | **98.8** | (+0.1) |
| | Proxy-anchor[†] [11] | BN | 512 | 78.9 | - | 90.6 | - | 96.1 | - | 98.5 | - |
| | *MemVir* + Proxy-anchor | BN | 512 | 79.7 | (+0.8) | **91.0** | (+0.4) | **96.3** | (+0.2) | 98.6 | (+0.1) |
| - | Average boost | - | - | - | (+1.1) | - | (+0.5) | - | (+0.2) | - | (+0.1) |
| | Minimum boost | - | - | - | (+0.8) | - | (+0.1) | - | (0.0) | - | (0.0) |
| | Maximum boost | - | - | - | (+1.5) | - | (+1.3) | - | (+0.6) | - | (+0.3) |

Table I. **[Conventional evaluation]** Recall@k (%) on Standford Online Products dataset in image retrieval task. Method type (T) is denoted by abbreviations (*Ens*: ensemble, *Gen*: sample generation, *M*: memory-based, *Pair*: pair-based losses, *Softmax variant / Proxy*: softmax variants and proxy-based losses). Backbone network (Net) also is denoted by abbreviations (*G*: GoogleNet [27], *BN*: BN-Inception [9]). [†] denotes evaluation in a fair setting described in Section C.2.1.

| SOP [21] | Concatenated (512-dim) | | | Separated (128-dim) | | |
|---|---|---|---|---|---|---|
| Loss | P@1 | RP | MAP@R | P@1 | RP | MAP@R |
| Norm-softmax [29] | 75.67 ± 0.17 | 50.01 ± 0.22 | 47.13 ± 0.22 | 71.65 ± 0.14 | 45.32 ± 0.17 | 42.35 ± 0.16 |
| *MemVir* + Norm-softmax | **75.77 ± 0.20** | **50.24 ± 0.22** | **47.45 ± 0.25** | **71.97 ± 0.15** | **45.47 ± 0.14** | **42.44 ± 0.14** |
| CosFace [30] | 75.79 ± 0.14 | 49.77 ± 0.19 | 46.92 ± 0.19 | **70.71 ± 0.19** | 43.56 ± 0.21 | 40.69 ± 0.21 |
| *MemVir* + CosFace | **75.88 ± 0.27** | **49.95 ± 0.37** | **47.18 ± 0.38** | 70.25 ± 0.18 | **43.82 ± 0.20** | **40.91 ± 0.19** |
| ArcFace [2] | **76.20 ± 0.27** | 50.27 ± 0.38 | 47.41 ± 0.40 | 70.88 ± 1.51 | 44.00 ± 1.26 | 41.11 ± 1.22 |
| *MemVir* + ArcFace | 76.05 ± 0.30 | **50.56 ± 0.33** | **47.75 ± 0.32** | **71.18 ± 0.27** | **44.31 ± 0.28** | **41.26 ± 0.28** |
| Proxy-NCA [19] | 75.89 ± 0.17 | 50.10 ± 0.22 | 47.22 ± 0.21 | 71.30 ± 0.20 | 44.71 ± 0.21 | 41.74 ± 0.21 |
| *MemVir* + Proxy-NCA | **76.97 ± 0.31** | **50.81 ± 0.26** | **48.02 ± 0.27** | **72.11 ± 0.25** | **44.82 ± 0.22** | **41.93 ± 0.18** |
| Proxy-ancho [11] | 75.37 ± 0.15 | 50.19 ± 0.14 | 47.25 ± 0.15 | 71.56 ± 0.11 | 46.13 ± 0.21 | 43.03 ± 0.21 |
| *MemVir* + Proxy-anchor | **77.80 ± 0.17** | **53.21 ± 0.12** | **50.35 ± 0.13** | **74.30 ± 0.18** | **48.94 ± 0.16** | **45.93 ± 0.15** |

Table J. **[MLRC evaluation]** Performance (%) on Standford Online Products dataset in image retrieval. We report the performance of concatenated 512-dim and separated 128-dim. Bold numbers indicate the best score within the same loss.



\end{document}